\documentclass[acmtog]{acmart}

\AtBeginDocument{%
  \providecommand\BibTeX{{%
    \normalfont B\kern-0.5em{\scshape i\kern-0.25em b}\kern-0.8em\TeX}}}

\definecolor{Red}{cmyk}{0,1,1,0}
\definecolor{Green}{cmyk}{1,0,1,0}
\definecolor{Cyan}{cmyk}{1,0,0,0}
\definecolor{Purple}{cmyk}{0.45,0.86,0,0}
\definecolor{Rosolic}{cmyk}{0.00,1.00,0.50,0}
\definecolor{Blue}{cmyk}{1.00,1.00,0.00,0}
\definecolor{BlueViolet}{cmyk}{0.86,0.91,0,0.04}
\definecolor{NavyBlue}{cmyk}{0.94,0.54,0,0}

\newcommand{\chen}[1]{{#1}}
\newcommand{\revised}[1]{{#1}}
\newcommand{\minor}[1]{{#1}}
\newcommand{\pang}[1]{{#1}}

\let\vec=\mathbf

\usepackage{amsmath}
\begin{document}

\title{TightCap: 3D Human Shape Capture with Clothing Tightness Field}


% \affiliation{%
% 	\institution{Institute for Clarity in Documentation}
% 	\streetaddress{P.O. Box 1212}
% 	\city{Dublin}
% 	\state{Ohio}
% 	\postcode{43017-6221}
% }

\author{Xin Chen*}
\affiliation{%
\institution{ShanghaiTech University, University of Chinese Academy of Sciences}
\country{China}}
\email{chenxin2@shanghaitech.edu.cn}

\author{Anqi Pang}
\authornote{The authors are contributed equally.}
\affiliation{%
\institution{ShanghaiTech University, University of Chinese Academy of Sciences}
\country{China}}
\email{pangaq@shanghaitech.edu.cn}

\author{Wei Yang}
\affiliation{%
\institution{School of Computer Science \& Technology, Huazhong University of Science \& Technology}
\country{China}}
\email{wyangcs@udel.edu}

\author{Peihao Wang}
\affiliation{%
\institution{ShanghaiTech University}
\country{China}}
\email{peihaowang@utexas.edu}

\author{Lan Xu}
\authornote{The corresponding authors.}
\affiliation{%
\institution{ShanghaiTech University}
\country{China}}
\email{xulan1@shanghaitech.edu.cn}

\author{Jingyi Yu$^{\dagger}$}
\affiliation{%
\institution{ShanghaiTech University}
\country{China}}
\email{yujingyi@shanghaitech.edu.cn}

\authorsaddresses{Authors’ addresses: Xin Chen, ShanghaiTech University, Shanghai Institute of Microsystem and Information Technology, Chinese Academy of Sciences, University of Chinese Academy of Sciences, China, chenxin2@shanghaitech.edu.cn; Anqi Pang, ShanghaiTech University, University of Chinese Academy of Sciences, China, pangaq@shanghaitech.edu.cn; Wei Yang, School of Computer Science \& Technology, Huazhong University of Science \& Technology, DGene Inc., China, wyangcs@udel.edu; Peihao Wang, ShanghaiTech University, China, peihaowang@utexas.edu; Lan Xu, ShanghaiTech University, China, xulan1@shanghaitech.edu.cn; Jingyi Yu, ShanghaiTech University, China, yujingyi@shanghaitech.edu.cn.}

% Some very useful LaTeX packages include:
% (uncomment the ones you want to load)
% \usepackage{times}
% \usepackage{epsfig}
% \usepackage{graphicx}
% \usepackage{amsmath}
% \usepackage{amssymb}
% \usepackage{booktabs}
% \usepackage{CJKutf8}
% \usepackage{caption}
% \usepackage{multirow}
% \usepackage{float}
%\usepackage[breaklinks=true,bookmarks=false]{hyperref}

\newcommand{\etal}{\textit{et al}.}
\newcommand{\tabincell}[2]{\begin{tabular}{@{}#1@{}}#2\end{tabular}}

\renewcommand{\shortauthors}{Trovato and Tobin, et al.}

\begin{abstract}
	% 1. Critical meaning
	In this paper, we present TightCap, a data-driven scheme to capture both the human shape and dressed garments accurately with only a single 3D human scan, which enables numerous applications such as virtual try-on, biometrics, and body evaluation.
	% 2. Key idea
	To break the severe variations of the human poses and garments, we propose to model the clothing tightness field – the displacements from the garments to the human shape implicitly in the global UV texturing domain.
	% 3. Key technical component
	To this end, we utilize an enhanced statistical human template and an effective multi-stage alignment scheme to map the 3D scan into a hybrid 2D geometry image. Based on this 2D representation, we propose a novel framework to predict clothing tightness field via a novel tightness formulation, as well as an effective optimization scheme to further reconstruct multi-layer human shape and garments under various clothing categories and human postures. We further propose a new clothing tightness dataset (CTD) of human scans with a large variety of clothing styles, poses, and corresponding ground-truth human shapes to stimulate further research.
	% 4. Superior experiment result
	Extensive experiments demonstrate the effectiveness of our TightCap to achieve the high-quality human shape and dressed garments reconstruction, as well as the further applications for clothing segmentation, retargeting, and animation.
\end{abstract}




\maketitle

%\begin{CJK}{UTF8}{gbsn}
%%%%%%%%%%%%%%%%%%%%%%%%%%%%%%%%%%%%%%%%%%%%%%%%%%%%%%%%%%%%%%%%%%%%%%%%%%%%%%%%%%%%%%%%%%%%%%%%%%%%%%%%%%%%%%%%%%%%%%%%
\section{Introduction}
\label{sec:introduction}
% 1. top view of the topic 
With the popularity of commodity 3D scanners such as Microsoft Kinect or ASUS Xtions, it has become increasingly common to create 3D human models in place of traditional 2D images.
How to further reconstruct the human shape as well as dressed garments for challenging human and clothing variations evolves as a cutting-edge technique requiring both refinement and robustness, which has attracted the attention of both the computer graphics and computer vision communities.

\begin{figure}[t]
    \centering
    \includegraphics[width=\linewidth]{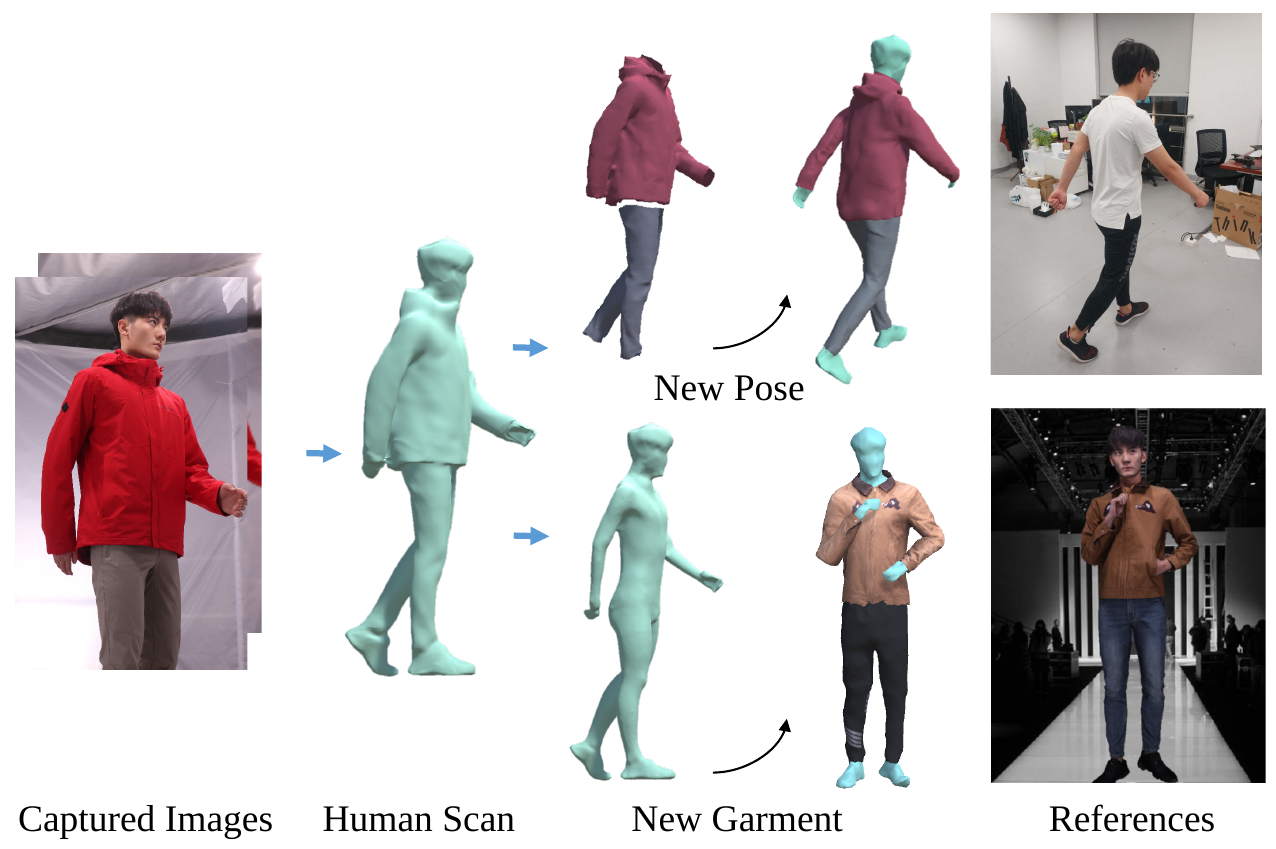}
    \caption{Our method first reconstructs a clothed 3D scan from images, and predicts the underlying human body shape as well as segments the garments. This illustrates how we can support a range of applications related to multi-layer avatar generation with detected pose (Right top) and try-on with pre-segmented garment models (Right down).
    }
    %\caption{Our method takes a clothed 3D scan (Left) as input and predicts the clothing tightness (Second) with the assistance of parametric human model and geometry images. With the predicted tightness, we can predict the underlying human body shape from the 3D scan as well as segment the clothes (Thrid). This illustrates how we can support a range of applications related to multi-layer avatar generation (Right), cloth segmentation, and try-on.}

    \label{fig:teaser}
\end{figure}

% 2. Our focus and problems of previous methods, NOTE: not deep enough
% 2.1 our focus
% \label{R4Q3_1}
\chen{Capturing the accurate human shape and garments respectively from only a single complete pre-scanned 3D mesh of a performer is essential for numerous applications (see Fig.~\ref{fig:teaser}) such as virtual try-on, biometrics, and body evaluation for gymnastics and sports.}
% \textcolor{blue}{[Ma: It is repeated to say "In this paper, we propose..." since it also occurs in the next paragraph. It is better to just throw the problem/task here, like "Capturing the accurate human shape and garments respectively from only a single pre-scanned clothed 3D mesh is essential for.... However, it is very challenging  because in reality,....."]}
% In this paper, we propose an approach for capturing the accurate human shape and the corresponding garments with only a single complete pre-scanned 3D mesh of the performer, which is essential for numerous applications (see Fig.~\ref{fig:teaser}) such as virtual try-on, biometrics, and body evaluation for gymnastics and sports.
%
% 2.2 general challenges
\chen{However, in reality, the human and clothing geometry can exhibit significant variations: borrowing jargon from clothing manufactures, clothing can be loose -- large clothing-body gaps to allow a full range motion, fitted -- a slimmer, athletic cut eliminating the bulk of the extra fabric, and compression -- ultra-tight, second-skin fit.
% As illustrated in Fig.~\ref{fig:smpl_fitting}
Such a variety of clothing categories and looseness makes it very challenging to capture the accurate human shape, let alone the variety of human postures and the further garment reconstruction.}
%
% 2.3 Mesh-based method: no cloth-tightness?
Earlier approaches~\citep{yang2016estimation, zhang2017detailed} utilize statistical body models, like SCAPE~\citep{SCPAE2005} and SMPL~\citep{loper2015smpl}, to optimize the human shape parameters in the model space or the vertex displacements from the human template to the captured 3D scans.
However, they are restricted by the generalization ability of the naked human template to handle various clothing geometry.
%Previous approaches have focused on approximating human shape from a single or multiple viewpoints, preferable with fitted or compression clothing. These approaches adopt optimization schemes to estimate the best model that fits the imagery, pose, and motion data, by assuming clothing a thin and fit layer over the skin. 
% 2.4 image-based method: only projection observation, low accuracy
% 
On the other hand, the learning-based techniques~\citep{kanazawa2018end, jiang2020bcnet, bhatnagar2019multi, alldieck19cvpr} have achieved significant progress recently to infer detailed geometry and body shape from images, but these image-based approaches suffer from scale ambiguity and inferior human shape accuracy.
%
% More importantly, these approaches above pay less attention to
% However, nearly all existing approaches conduct reconstruction without considering the effects of clothing, or more precisely, the tightness of clothing. 

% 3. Key idea of our approach
\chen{In this paper, we propose TightCap, a data-driven approach to capture both human shape and dressed garments robustly with only a single complete 3D scanned mesh of the performer, which can be generated from multi-view RGB cameras~\cite{schonberger2016pixelwise} or a single depth camera~\cite{newcombe2015CVPR}. And, we outperform existing state-of-the-art methods to significantly produce 5.7 millimeters accuracy.}
% \textcolor{blue}{[Ma: to outperform? Maybe you can use a single sentence to stress your performance.]}
%
Our novel scheme introduces the clothing tightness field which represents the displacements from various garments to the underlying human body model in a global UV texturing domain, so as to implicitly model the influence of various clothing categories, looseness, and human postures.

%Fig.\ref{fig:teaser} shows our estimated tightness, underlying body shape with its clothing of a human.
%
% We observe that humans can quickly identify clothing tightness (loose vs. fit vs. compression) as important prior to shape estimation and seek to develop a similar learning-based pipeline. 
%
% Specifically, we set out to combine global and local inferences: the former includes clothing styles and types, and the latter includes shape deformations such as folds and puffiness.

% 4. Our technical Pipeline illustration
% template alignment
% tightness learning
% body shape estimation
More specifically, to estimate the clothing tightness field for various garment categories and human postures in a data-driven manner, we first map the captured human 3D scan into a global geometry image~\citep{gu2002geometry}, called clothed-GI.
To this end, we extend the statistical human model SMPL~\citep{loper2015smpl} by subdividing its geometry features around the garment boundary (e.g., neck, wrist, waist, ankle, and etc.) for the generalization to various clothing categories.
Then, an effective multi-stage alignment scheme is adopted to warp the enhanced template to the captured scan, which jointly leverages the skeleton, silhouette, and geometry information in a coarse-to-fine manner to handle human pose and garment variations.
Second, we generate a hybrid feature embedding from the generated clothed-GI, including per-pixel texture, position, and normal.
We further utilize a conditional generative adversarial network (GAN) to regress per-pixel clothing tightness in the UV texturing domain to handle human garment and posture variations, with the aid of a novel per-vertex tightness formulation and a new 3D dataset that consists of a large variety of clothing including T and long shirt, short/long/down coat, hooded jacket, pants, skirt/dress, and the corresponding 3D human shapes.
% (R1) What is "modified" in the conditional GAN? Simply changing the number of inputs and outputs without fundamental modification is considered as the original work.
%
Finally, we propose an effective optimization scheme to reconstruct both the inner human shape and the multi-layer dressed garments accurately from the predicted tightness map in the geometry image domain.
Comprehensive experiments on both public and our captured datasets show that, compared with the state-of-the-art, with only a single captured 3D scan, our approach significantly improves the accuracy of human shape prediction, especially under various loose and fitted clothing.
We further demonstrate how the recovered multi-layer human geometry can be applied to automatically segment clothing from the human body on 3D meshes as well as cloth retargeting and animation.
% 5. More specific technical contributions
% 1) A novel data-driven body shape estimation scheme, which is the first to model the cloth tightness
% 2) A robust multi-stage alignment approach to generate GI(quite misleading)?
% 3) A novel tightness learning scheme based on a novel tightness formulation as well as an effective body and clothing recovery scheme
To summarize, our main contributions include:
\begin{itemize}
    \setlength\itemsep{0em}
    \item A novel and superior human shape and garment capture scheme with a single captured 3D scan, which models clothing tightness field to handle the garment and posture variations implicitly in the UV texturing domain.
    \item An effective multi-stage alignment approach to enable clothed-GI generation from the captured scan with the aid of an enhanced statistical model.
    \item An novel tightness map learning scheme based on a novel per-vertex tightness formulation as well as an effective optimization scheme to recover both the human shape and garments.
    \item To stimulate further research, we make available our clothing tightness dataset (CTD) of totaling 880 human models with 228 different garments under various human postures as well as the ground truth human shapes.
\end{itemize}

\section{Related Work}
% (R1）organize related work, contrast the proposed method with the existing methods
% 增加
\paragraph{Human and Garment Modeling.}
Most of the early works on human modeling can be categorized as multi-view stereo (MVS) vs. depth fusion-based approaches. The former approaches~\cite{Furukawa2013, Strecha2008,Newcombe2011,collet2015high} employ correspondence matching and triangulation. For example, \citet{collet2015high} use a dense set of RGB and IR video cameras, producing high-quality 3D human results.
The latter approaches~\citep{Bogo2015detailed, newcombe2015CVPR, Yu2017ICCV, Dou2016TOG, 8708933, li2020robust,pang2021few,suo2021neuralhumanfvv} use active sensors such as structured light and Time-of-Flight (ToF) range scanning (e.g., Microsoft Kinect I and II, respectively), which have a much lower cost. For example, DynamicFusion~\citep{newcombe2015CVPR} compensates for geometric changes due to motion captured from a single RGB-D sensor.
UnstructuredFusion~\citep{8708933} and RoubustFusion~\citep{su2020robustfusion} utilizes serveral RGBD cameras to capture textured 4D human scans.
Recently, many learning based works utilize statistical body models, like SMPL~\citep{loper2015smpl}, to capture/reconstruct human with clothing ~\citep{alldieck19cvpr, zheng2019deephuman, bhatnagar2019multi, bhatnagar2019mgn} or recover human body~\citep{Pavlakos2018CVPR, kanazawa2018end, Challencap2021, chen2021sportscap} with the 2D keypoint detectors \citep{cao2018openpose,li2019pose2body}. Also, some notable works~\citep{Joo2019studio, Joo2018total, Xiang2019CVPR}, from Carnegie Mellon University, capture single or multiple 3D humans from 3D pose and body shape based on the multi-view panoptic studio. Similar to their tasks, \citet{SMPLX2019} can also capture face, hand, body and expression with the SMPLX from a single image.
For other 3D representations of clothed human, \chen{SiCloPe~\cite{natsume2019siclope} utilizes silhouette-based representation for modeling clothed human bodies using deep generative models.
PIFu~\citep{saito2019pifu} proposes a pixel-aligned implicit representation to digitize detailed clothed humans from images, and PIFuHD~\citep{saito2020pifuhd} formulate a multi-level architecture to address the memory limitations of the hardware.
PIFusion~\citep{li2020robust} combines learning-based 3D recovery with volumetric non-rigid fusion to generate clothed human scans.
\citet{bhatnagar2020ipnet} combines implicit functions and parametric representations to reconstruct 3D models of people.
Most of these reconstruction works focus on only one of two layers, the top surface layer of human scan or the skin layer of unclothed body shape. In TightCap, we model both the cloth and body layer with tightness field, \minor{a specific displacement from the cloth to the body}, and build a data-driven method to estimate body shape and build a multi-layer avatar.}

%\paragraph{Garment Modeling.}
Garment modeling could be included in general human modeling, as we introduced, but it usually assumes the clothes and body skin belong to the same surface layer, like \citet{alldieck19cvpr, alldieck2018video, natsume2019siclope, pumarola20193dpeople, saito2019pifu, lazova2019360, alldieck2019tex2shape, saito2020pifuhd}.
\revised{Different from these methods, some other works \citep{neophytou2014layered, pons2017clothcap, yu2019simulcap} propose the idea of the multi-layer human model for garment modeling.
DoubleFusion~\citep{2018DoubleFusion} presents a system to reconstruct cloth geometry and inner body shape based on the parametric body model. Their approach allows the subject to wear casual clothing and separately treats the inner body shape and outer clothing geometry.
ClothCap \citep{pons2017clothcap} and SimulCap \citep{yu2019simulcap} also use SMPL as template model to help model garment from reconstructed human. 
For learning-based methods, DeepWrinkles \citep{Zorah2018ECCV} proposes a data-driven framework to estimate garment wrinkles from the body motion. Moreover, CAPE~\citep{ma20autoenclother} and TailorNet~\citep{patel20tailornet} use learning-based methods to generate a 3D mesh model of clothed people or directly proposes a neural garment model with pose and shape.
ARCH~\citep{huang2020arch} proposes a pose-aware model that produces 3D rigged clothed human avatars from a single image. 
\citet{tiwari20sizer} propose a dataset of people with clothing size variation and model 3D clothing conditioned on body shape and garment size parameters. \label{R5Q8_2}
Most mentioned works of garment modeling, such as \citet{lazova2019360, alldieck2019tex2shape, pumarola20193dpeople}\label{R5Q8}, utilize geometric image representation and the parametric body shape as the prior to reconstruct clothed human shape. Contrary to their tasks, we utilize the reconstruction results of clothed human, and focus on predicting the tightness map of different types of clothing. And, our approach recovers both the personalized body shape under clothing and the reliable multi-layer avatar.}

\paragraph{Shape Under Clothing.}
\chen{Estimating body shape under clothing is more challenging, because the clothing occludes the original body shape. 
%\textcolor{blue}{[Ma: followed by a short reason]}.
Earlier methods \citep{Balan2008ECCV, zhang2017detailed, pons2017clothcap} employ a statistical or parametric 3D body model, like SCAPE \cite{SCPAE2005} and SMPL \cite{loper2015smpl}. \citep{Balan2008ECCV} build on the concept of the visual hull under the assumption that the clothing becomes looser or tighter on different body parts as a person moves.
% They estimate a maximal silhouette-consistent parametric shape (MSCPS) from several images of a person with both minimal and normal clothing.
\citet{Wuhrer2014CVIU} estimate body shape from static scans or motion sequences by modeling body shape variation with a skeleton-based deformation.
Other approaches~ \citep{Hasler2009, zhang2017detailed, pons2017clothcap, 2018DoubleFusion} utilize the parametric body model as the prior of shape, and attempt to optimize the body shape with the boundary constraint of the clothed human scans. ClothCap~\citep{pons2017clothcap} utilizes a multi-part 3D model to estimate a minimally clothed shape under the clothing and tracks the deformations.
DoubleFusion \citep{2018DoubleFusion} use SMPL to estimate body shape as one layer of their double-layer model.
\citet{Yang2018ECCV} propose a statistical regression model for the variability of the clothing for capturing underlying shapes.
However, they require the subject to wear ultra-tight or fitted clothing and only focus on several types of clothing. Human body shape estimation in wide and puffy clothing is significantly more difficult than in fitted clothing. Therefore, our approach is different from these approaches. We not only utilize a prior of template body mesh, but also exploit a data-driven manner for predicting the inner body shape under hundreds of various clothes.}
For learning-based methods, \citet{Leonid2016CVPR, Wei2016CVPR, Newell2016ECCV} learn articulated body poses of humans from their occluded body parts via convolutional networks (ConvNets). \citet{Lassner2017CVPR} predict body segments and landmarks from annotated human pose datasets, and conducts body pose estimation with clothing and 3D body fitting. \citet{Lassner2017ICCV} present a generative model of the full body in clothing, but focusing more on appearance generation than body shape estimation. Especially, HMR~\citep{kanazawa2018end} proposes an end-to-end ConvNet to recover the parameters of SMPL, for generating a 3D human body mesh from a single image. \citet{Pavlakos2018CVPR} refine this similar generated body mesh by projecting body shape back to the 2D image for full-body pose and shape estimation. Their techniques rely on parameter prediction from the body model and body pose accuracy. Different from these works which only capturing the pose and shape parameters from images, our approach models both clothing and body. The clothing layer produces significant help for more accurate and realistic results.

\begin{figure*}[!htb]
    \centering
    \includegraphics[width=0.98\linewidth]{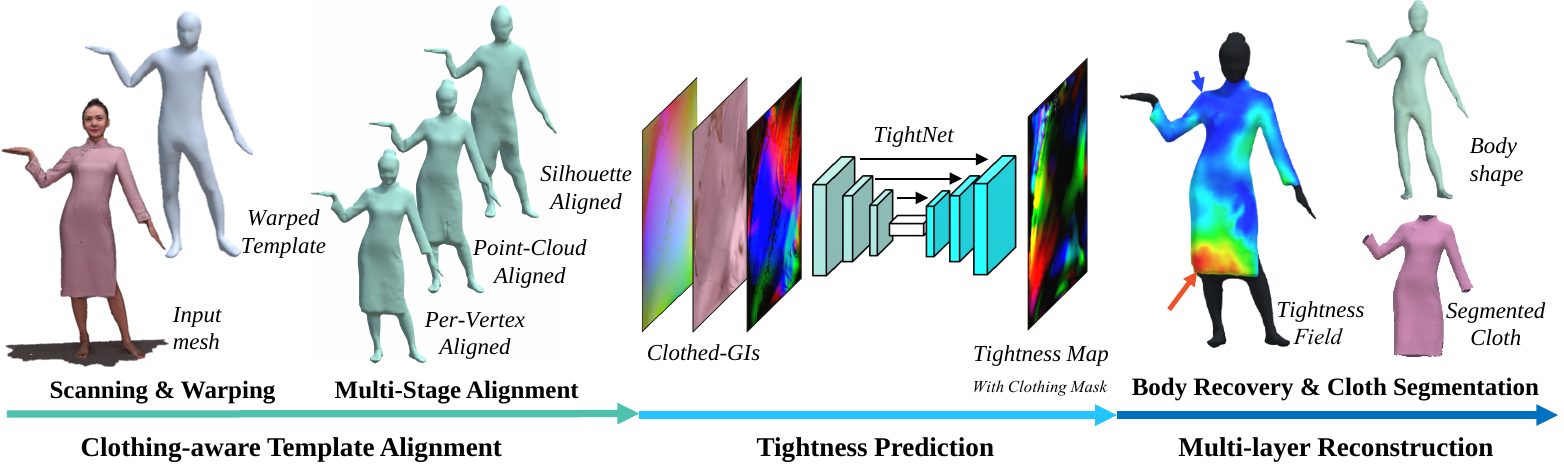}
    \caption{The pipeline of TightCap. The first step is to warp our enhanced clothing-adapted SMPL with scanned mesh. Then, we deform warped mesh using \emph{Multi-Stage Alignment}. Next, we estimate the tightness map and the clothing mask from mapped clothed-GI with \emph{Tightness Prediction}. The final step, \emph{Multi-layer Reconstruction} is to recover body shape from the predicted tightness field on the mesh, and segment cloth.}
    \label{fig:pipeline}
\end{figure*}

\label{related_work_comp}
\chen{To contrast our approach with most related methods of shape estimation, we summarize these works in three different technical schemes, including the shape optimization under the geometric constrain of clothing, the learning-based body shape recovery from a single image, the displacement-based human capture in map or mesh (per-vertex). For geometric optimization, these methods, such as \citet{Yang2018ECCV} and \citet{zhang2017detailed}, focus on optimizing the variations of the body with geometric constraints of clothing. Different from them, our approach builds a learning-based technique from different subjects in hundreds of different clothes, not only predicting the tightness field between the human body and the cloth layer but also segmenting clothing. With only a static input mesh rather than mesh sequence, our approach can leverage more commercial 3D sensors, such as ToF cameras on mobile phones, to support mobile virtual fitting tasks. For learning-based body shape recovery, like HMR~\citep{kanazawa2018end} and SMPL-X~\citep{SMPLX2019} (based on learning-based pose estimator), our method can produce more reliable and accurate bodies with the help of the reconstruction of the clothing layer. For these methods based on displacement map~\citep{lazova2019360, pumarola20193dpeople, alldieck2019tex2shape} or per-vertex displacement~\citep{alldieck19cvpr, alldieck2018video}, most of these works focus on capturing a clothed human with the template model, rather than focusing on the personalized shape estimation. Their displacement/offset usually points from an impersonalized template body to the cloth, ignoring the influence of various clothing types on body shape. Instead, we consider the tightness field for shape recovery under hundreds of different clothes. Moreover, with the prediction of the tightness map and clothing mask, our approach can easily separate the clothing from the personalized body shape, benefiting many applications like virtual fitting and size measurement.}
%%%%%%%%%%%%%%%%%%%%%%%%%%%%%%%%%%%%%%%%%%%%%%%%%%%%%%%%%%%%%%%%%%%%%%%%%%%%%%%%%%%%%%%%%%%%%%%%%%%%%%%%%%%%%%%%%%%%%%%%
\section{Overview}
Our goal in this paper is to reconstruct the human shape and corresponding garments accurately with a single captured 3D scan.
To handle this challenging problem under the huge variations of clothing categories and human postures, we utilize the clothing tightness field in a data-driven manner, which relies on a representative dataset with both clothed human scans and corresponding inner body shapes.
Thus, we collect a new clothing tightness dataset (CTD) with dressed human meshes under various clothing and poses, which are reconstructed via a dome system equipped with 80 RGB cameras using the MVS approach~\citep{schonberger2016pixelwise} or an RGB-D sensor using DynamicFusion~\citep{newcombe2015CVPR}.
The corresponding ground truth naked human shapes are obtained via the same system and further manually re-posed and annotated by multiple professional artists.

\label{R4Q1}
\chen{Different from the reconstruction of clothed human. our system focus on learning personalized body shapes and multi-layer avatars from various clothed human scans. To formulate this system around the learning-based module, we build three modules: Template Alignment, Tightness Prediction, and Multi-layer Reconstruction.} Fig.~\ref{fig:pipeline} illustrates these three high-level components of the algorithm pipeline, which achieves considerably more accurate body shape reconstruction results than previous methods. 

% enhanced SMPL; multi-stage alignment
\paragraph{Human Template Alignment.}
\chen{To model the garment and posture variations implicitly with a learning-based approach, template alignment is an essential process for most network models. Considering the GAN-based CNNs on the image domain can effectively generate more high-frequency details of clothes than GCN with less memory in current hardware, a novel template alignment scheme is adopted to map the input 3D scan into the 2D clothed-GI.
To align various clothes with the template model, our scheme relies on a garment-specific human template extended (Sec.\ref{TemplateModel}) from the statistical model SMPL~\citep{loper2015smpl} with both pose parameters and embedded nodes.
And, we utilize a multi-stage alignment with different sensitivities of parameters, including joints, embedded nodes, and vertices, which jointly leverage the skeleton, silhouette, and geometry information in a coarse-to-fine manner (see Sec.\ref{alignment}). After alignment, we utilize geometry image mapping to unwrap the aligned surface into 2D clothed-GI (Sec.\ref{GeoImgRep}).}
%
%To this end, we extend the statistical human model SMPL~\citep{loper2015smpl} by subdividing its geometry features around garment boundary (e.g. neck, wrist, waist, ankle, etc.) for the generalization to various clothing categories.
%

\paragraph{Tightness Prediction.}
Based on the hybrid feature map from the above clothed-GI, we propose to predict the corresponding 2D clothing tightness map in a data-driven manner, which utilizes a novel per-vertex tightness formulation (see Sec.\ref{tightnessMeasure}).
\chen{While GAN-based networks \citep{lazova2019360, pumarola20193dpeople, alldieck2019tex2shape} achieve reliable results on reconstructed clothed human, we follow \minor{this GAN-based framework} to predict the tightness pointing from the clothing layer to the body layer for personalized body shape estimation}, and build a most effective learning framework based on conditional GAN, named TightNet (see Sec.\ref{TightNet}).

\paragraph{Multi-layer Reconstruction.}
Finally, we utilize the predicted tightness map to reconstruct the multi-layer human shape and dressed garments accurately via an optimization scheme based on Gaussian kernels (see Sec.\ref{ShapRecovery}). \chen{This optimization can produce more reliable and stable results with the prior of template shape, and help to correct the 3D artifacts for challenging cases.
% like crotch and oxter.
Such multi-layer reconstruction results further enable various applications such as immersive cloth retargeting and avatar animation.
%
%At the end, we present an exemplary application of our approach for avatar animation, which retargets the segmented clothing layer to a new body shape under a different pose.
%
The following sections provide the details of the full system.}
%we first prepare a dataset with both clothed human mesh and underlying body shape. %We collect raw clothed human mesh with the multi-camera dome system, as described in Sec.\ref{input3DMesh}. The ground-truth unclothed body shape is generated by artists (Sec.\ref{dataset}).

%%%%%%%%%%%%%%%%%%%%%%%%%%%%%%%%%%%%%%%%%%%%%%%%%%%%%%%%%%%%%%%%%%%%%%%%%%%%%%%%%%%%%%%%%%%%%%%%%%%%%%%%%%%%%%%%%%%%%%%%
\section{Clothed Human Template Alignment}
% 解释说明 为什么要用这样的方法 
% 1 没有办法直接获得corresponding 人工工作量太大了 
% 2 没有ground truth 也可以参考的数据集来辅助进行learning-based方法
%   我们的目标是 解决大规模 不同样式衣服的aligment的问题 只有人裸体和少部分穿衣服align的工作 比如FAUST, 其中最新的SIZER是解决xxx问题 也无法使用于这种alignment
% 3 也没有办法采用 事先做mark 因为要保留衣服的颜色
%   所以我们采用modify template + defomration-based alignment算法 
%   试图生成足够解决的 aligned mesh
%   具体 template的设计   具体 alignment算法的设计
\label{TemplateAlignment}
% Our scheme relies on a  human template extended from the statistical model SMPL~\citep{loper2015smpl} (Sec.\ref{TemplateModel}), and a multi-stage alignment which jointly leverages the skeleton, silhouette and geometry information in a coarse-to-fine manner (see Sec.\ref{GeometryImages}).
Under the canonical pose, the tightness field between various clothing and body shapes share a similar distribution with the same direction of gravity, e.g., the clothes tend to be loose around the human oxter and crotch, and tight on the shoulder and chest, 
% \textcolor{blue}{[Ma: Is the description true for most cases? How could you get the rules?]}
which implies that the tightness field of clothes with the underlying human body can be predicted in a data-driven manner. 
To this end, we utilize a garment-specific statistical human model on top of SMPL~\citep{loper2015smpl} (Sec. \ref{TemplateModel}), and adopt an effective multi-stage alignment to warp the enhanced template to the captured scan, which jointly leverages the skeleton, silhouette, and geometry information in a coarse-to-fine manner to handle human pose and garment variations (Sec. \ref{alignment}). 
We further transfer the input 3D human scan into a global and regular geometric image~\citep{gu2002geometry}, called clothed-GI (Sec. \ref{GeoImgRep}), to maintain the continuity of semantic human body distribution and implicitly model the variations of garments and human postures for the clothing tightness training and prediction.
We provide the details of each design in the following.

\subsection{Clothing Adapted Human Template Model}\label{TemplateModel}
\begin{figure*}[t!]
	\centering
	\includegraphics[width= \linewidth]{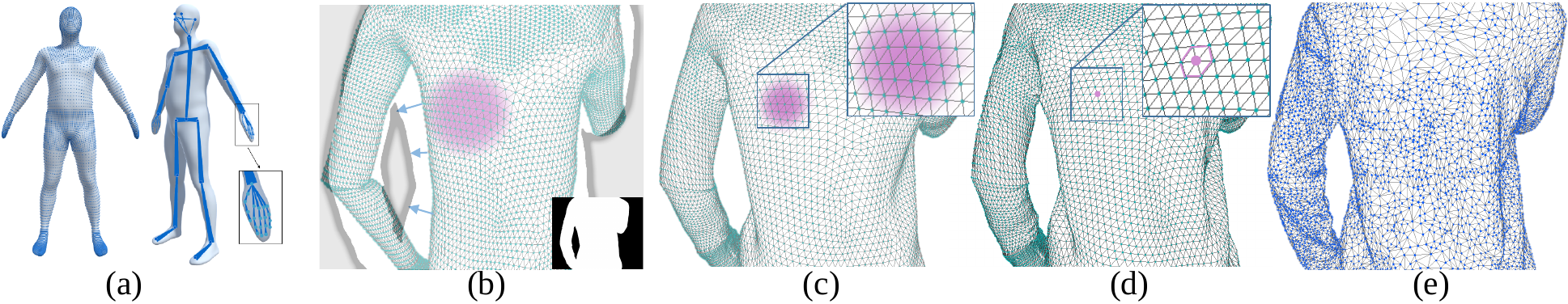}
	\caption{Each stage of our alignment approach. (a) The template model for alignment. (b) The first stage, silhouette based deformation. (c) The second stage, point cloud based deformation. (d) The third stage, per-vertex deformation. (e) The referenced mesh (target mesh). The pink regions indicate the different ranges of ED nodes, while the wires are topological structure. } % \textcolor{blue}{[Ma: What do pink areas, points and circles mean in this figure?]}
	\label{fig:CA-SMPL}
	\label{fig:3_stage}
\end{figure*}

\chen{The most successful parametric human models, i.e., the SMPL~\citep{loper2015smpl} and SMPL-X\citep{SMPLX2019}, focus on modeling the naked human body with various poses and shapes, while the displacement-based vertices movement produces the generalization to represent various clothes. To suit most clothed human meshes on shoes and with closed or even hidden hands in our dataset, we modify the human model SMPL~\citep{loper2015smpl}. We simplify face/hands/feet and subdividing its geometry features around the garment boundary (e.g., neck, wrist, waist, ankle.) to generalize these various clothing categories.}
% \textcolor{blue}{[Ma: Why utilize semantic density distribution?]}}
% \chen{which suffer from limited generation ability to handle clothing variations. xxxxxxxxxxxxxxxxxxxxxxxxxxxxxxxxxxxxxxxxxx}
% as demonstrated in Fig.~\ref{fig:smpl_fitting}
We also simplify the template mesh around the ears, nose, and fingers for efficiency and rig the modified model with the skeleton defined by OpenPose~\citep{cao2017realtime, simon2017hand}, i.e., 23 joints for the main body part and 21 joints for each hand, as shown in Fig.\ref{fig:CA-SMPL} (a).
The utilized clothing-adapted SMPL (CA-SMPL) model, denoted as $\mathcal{M}_T$, contains $N_{\mathbf{M}} = 14985$ vertices, $N_{\mathbf{F}} = 29966$ facets and $N_{\mathbf{J}} = 65$ joints, which is summarized as follows:
\begin{align}
    \mathcal{M}_\text{T} = \left\{\mathbf{M} \in \mathbb{R}^{N_{\mathbf{M}} \times 3}, \mathbf{F} \in  \mathbb{R}^{N_{\mathbf{F}} \times 3}, \mathbf{J} \in \mathbb{R}^{N_{\mathbf{J}} \times 4 + 3} \right\},
    \label{eq0}
\end{align}
where $\mathbf{M}$, $\mathbf{F}$ and $\mathbf{J}$ denote the parameter sets of vertices, facets and joints, respectively.
% is the $3 \times 3$ diagonal matrix with the $X, Y, Z$  on the diagonal. The vertices of the hand in world
\chen{Different from the original SMPL model using both pose and shape parameters $\vec{\beta}$ and $\vec{\theta}$, we utilize the scale of bone $\mathbf{S}$ to produce personalized lengths of bone without the shape parameters, which is similar to the scaling factors $\phi_{j}$ in \cite{Joo2018total}. We only use the pose parameters to drive the template, thus do not need the joint regressor $J(\vec{\beta} ; \mathcal{J}, \overline{\mathbf{T}}, \mathcal{S})$ in SMPL.}
Hence, the utilized pose parameters in the adopted human template are as follows:
\begin{align}
    \mathbf{J} = \left\{ \mathbf{\Theta} \in \mathbb{R}^{N_{\mathbf{J}} \times 3}, \mathbf{S} \in \mathbb{R}^{N_{\mathbf{J}}}, \mathbf{m} \in \mathbb{R}^3 \right\},
\end{align}
including the joint angles $\mathbf{\Theta}$ with axis-angle representation, the scaling factors $\mathbf{S}$ of each joint along the bone direction, and the global translation $\mathbf{m}$. 
% (R1) Overall the formulation is slightly deviated from SMPL (for example, how are the joints determined? Also why is per-joint scale necessary? In SMPL, joints are obtained by regressing from the surface body mesh.). Any necessary modifications need to be mentioned with motivation and original formulation of SMPL for better contrast.
%
Furthermore, let $M(\hat{\mathbf{J}})$ denote the skeletal motion of the human template after applied the joint parameters $\hat{\mathbf{J}}$, and $\hat{\mathbf{M}}$ represents the warped vertices of the template.

% ED-motion
To enable robust alignment, we utilize the embedded deformation (ED)~\citep{sumner2007embedded} by sampling the ED graph on the above enhanced human template, which is formulated as follows:
\begin{align}
    \begin{split}
        \mathbf{\mathcal{G}} = \left\{ \mathbf{R} \in \mathbb{R}^{N_{\mathbf{G}} \times 3}, \mathbf{t} \in \mathbb{R}^{N_{\mathbf{G}} \times 3} \right\},
    \end{split}
    \label{EDGraph}
\end{align}
where $N_{\mathbf{G}}$ is the number of nodes in ED graph.
Then, the warping function $G_k$ of the $k$-th node applied to a vertex $\mathbf{v}$ consists of the rotation $\mathbf{R}_k \in \mathbf{SO}(3)$ and the translate $\mathbf{t_k \in \mathbb{R}^3}$, which is formulated as:
\begin{align}
    G_k(\mathbf{v}) = \mathbf{R}_k(\mathbf{v} - \hat{\mathbf{g}}_k) + \hat{\mathbf{g}}_k + \mathbf{t}_k,
\end{align}
where $\hat{\mathbf{g}}_k \in \mathbb{R}^3$ indicates the canonical position of the $k$-th node.
Thus, the $i$-th vertex $\mathbf{v}_i, i\in [1, N_{\mathbf{M}}]$ on the human template after applied the ED motion $\mathbf{\mathcal{G}}$ is formulated as:
\begin{align}
    \mathbf{v}_i(\mathcal{G}) = G(\hat{\mathbf{v}}_i) = \sum_{k \in N_\mathbf{G}} \mathbf{w}_{i,k}^{\mathbf{G}}G_k(\hat{\mathbf{v}}_i).
\end{align}
Here $\hat{\mathbf{v}}_i$ is the canonical position of vertex $i$ and $\mathbf{w}_{i,k}^{\mathbf{G}}$ is the skinning weight between the $i$-th vertex and the $k$-th ED node according to the Euclidean distance.
Please kindly refer to \citet{sumner2007embedded} for more details about the setting of skinning weight.

%Both MVS method or scan-based method still have a big challenge on these small parts on human body. The missing finger or complicated ear always leads to bad results of convergence. Hence, different from general human body mesh alignment, our approach focuses on clothing parts like wrinkles and clothing boundary, rather than body parts.

%\paragraph{Input 3D Mesh Acquisition.}
\chen{
\subsection{Human Scan Reconstruction.} \label{dataset_creation}
Our raw 3D human scans can be captured using a multi-view dome system or a single-view depth sensor.
For the former setting, we reconstruct using the MVS approach~\citep{schonberger2016pixelwise} from multi-view human images captured by a dome system equipped with 80 cameras. We also estimate the 2D human joints as in \citet{cao2017realtime, simon2017hand} for the 80 original views and the 30 synthetic views (see Sec. \ref{30_Cams}) in turn. While the 80 original views help to locate the 30 synthetic views for accurate 2D joint estimation, we obtain the 3D joints through triangulation as in~\citet{triggs1999bundle} for initialization.
The other setting is to reconstruct non-rigidly deforming human from commodity sensors, like ToF cameras on mobile phones and Kinect. We implement DynamicFusion~\citep{newcombe2015CVPR} as our reconstruction method to acquire dynamic human scans. We also estimate the 2D human joints for this single view and project it to the depth map as our rough 3D joints.
% 
%We warp the CA-SMPL $\mathcal{M}_T$ with rough 3D joints $\mathbf{J}_{mv}$ and compute vertex accordingly $\mathbf{M}_{warp} = M(\mathbf{J}_{mv})$.
% Start with $\mathbf{M}_{warp}$, our multi-stage deformation scheme works as follows.

\chen{We reconstruct most of our clothed human mesh sequences in our dataset with the multi-view system, while a small part of them are from the single-view depth system. The main reason is that the multi-view system can capture more realistic clothing movement rather than a single-view system without the details on the back. Meanwhile, we utilize the depth sensor to capture the dark color clothing without a rich image feature. For more generalization, we also utilize many synthetic clothed avatars from Adobe Fuse CC. Although these meshes can not provide realistic motion of body and clothing, the various color of clothing and skin are convenient for data augmentation. For more detail of our dataset, please refer to Sec. \ref{dataset}.}

Our multi-stage deformation scheme can adapt to different application settings. More technical details for each stage are provided as follows.

% For prepossessing, we warp the CA-SMPL $\mathcal{M}_T$ with the triangulated 3D joints $\mathbf{J}_{mv}$ and compute vertex accordingly $\mathbf{M}_{warp} = M(\mathbf{J}_{mv})$.
% Start with $\mathbf{M}_{warp}$, our multi-stage deformation scheme works as follows.

% Input 3D Mesh Acquisition
%The input of our method is raw 3D human meshes, which we reconstruct using MVS approach~\citep{schonberger2016pixelwise} from multi-view human images captured by a dome system equipped with 80 cameras.
% We also estimate the 2D human joints for each image as in \citet{cao2017realtime,simon2017hand} and obtain the 3D joints through triangulation as in~\citet{triggs1999bundle}.
% For prepossessing, we warp the CA-SMPL $\mathcal{M}_T$ with the triangulated 3D joints $\mathbf{J}_{mv}$ and compute vertex accordingly $\mathbf{M}_{warp} = M(\mathbf{J}_{mv})$.
% Start with $\mathbf{M}_{warp}$, our multi-stage deformation scheme works as follows.
}

\subsection{Multi-Stage Alignment} \label{alignment}
\label{R4Q1_2}
\chen{Both two popular learning-based approaches, the standard CNNs (image-domain) and the GCN (mesh-domain), need the ground truth of registered meshes. Considering the lack of existing large datasets of registered meshes, especially for the generalization to these various clothes in our dataset, the data-driven mesh alignment approaches are not suitable. Meanwhile, the appearance of clothes also need be maintained as one important feature for generalization. Thus, we avoid manually mark the points/patterns on the clothes for the correspondence, which is used in the registration of naked human meshes in FAUST \citep{Bogo:CVPR:2014}. To this end, we adopt a novel multi-stage alignment scheme to transfer the input 3D human scan into the consistent human template, which shares the same topology for various garment categories and human postures. 
% note: (R1) the system actually convert 3D mesh with the consistent topology. So "irregular" is probably not appropriate to describe 3D meshes.
%
Note that our scheme consists of the silhouette based, point cloud based and per-vertex deformation stages to optimize the non-rigid motions from the enhanced human template to the input 3D scan in a coarse-to-fine manner, as illustrated in Fig.~\ref{fig:3_stage}.}

\paragraph{Silhouette based deformation.}
We first align the enhanced template with the silhouette information from the captured scan using a coarse ED-graph to handle error-prone places due to holes or noise on the raw human scan.
\label{30_Cams}To fetch the silhouette information, we utilize a virtual capture system with $N_{\mathcal{C}}= 30$ synthetic cameras to view different areas of the captured 3D mesh.
Note that for capturing the neck, ankles, and wrists, we set two synthetic cameras orthogonal to each other.
Besides, five cameras with different view angles are arranged to capture the upper and lower body torso, respectively.
The resulting synthetic camera setting is formulated as follows:
\begin{align}
    \mathcal{C} = \left\{ \left( \mathbf{c}_j \in \mathbb{R}^6, w_j^\mathcal{C} \in \mathbb{R}^1 \right) \arrowvert j \in [0, N_{\mathcal{C}}) \right\},
    \label{eq2}
\end{align}
where $\mathbf{c}_j$ denotes extrinsic parameters of a camera and $w_j^\mathcal{C} \in [0.5, 1]$ represents the weighting factor for two different camera positions ( 0.5 for the torso regions and 1 for capturing limbs).
Such a semantic weighting strategy further improves the alignment results, especially for those boundary regions.
%
%We also estimate the 2D joints for the virtual rendered image using the 2D pose detection methods~\citep{cao2017realtime,simon2017hand} and obtain the 3D joints through triangulation as in~\citet{triggs1999bundle}.
We first warp the original human template model $\mathcal{M}_T$ with the rough 3D joints $\mathbf{J}_{mv}$ as our initial mesh for the following alignment.
Inspiring by previous silhouette deformation method~\citep{xu2018monoperfcap}, we render the high-resolution silhouette masks of the captured scan for all the virtual views and
phrase the coarse-level alignment by solving the following non-linear least-squares optimization problem:
\begin{align}
    E_{\text{S}}(\mathcal{G}) = E_{\text{mv}}^{\text{S}}(\mathcal{{G}}) + \lambda_{\text{reg}}^{\text{S}}E_{\text{reg}}^{\text{S}}(\mathcal{G}).
    \label{eq3}
\end{align}
Similar to \citet{xu2018monoperfcap}, our multi-view silhouette based data term $E_{\text{mv}}^{\text{S}}$ measures the 2D point-to-plane misalignment:
\begin{align}
    E_{\text{mv}}^{\text{S}}(\mathcal{G}) = \sum_{j \in \mathcal{C}} \frac{w_j^\mathcal{C}}{\lvert \mathbf{v}_j^{\text{S}} \rvert} \sum_{k \in \mathbf{v}_j^{\text{S}}} \Vert \mathbf{n}_k^{\mathrm{T}} \cdot (P_j(\mathbf{v}_i(\mathcal{G})) - \mathbf{p}_k) \Vert ^ 2 _2,
\end{align}
where $\mathbf{v}_j^{\text{S}}$ is the vertex set of virtual silhouettes of the input scan and $P_j(\cdot)$ is the projection function of the $j$-th camera.
For each silhouette point $\mathbf{p}_k \in \mathbb{R}^2$ with the 2D normal $\mathbf{n}_k \in \mathbb{R}^2$, we search its corresponding deformed vertex in the utilized human template, denoted as $\textbf{v}_i$, found via a projective look-up method in an Iterative Closest Point (ICP) manner.

Similar to \citet{sorkine2007rigid}, the regularity term $E_{\text{reg}}^{\text{S}}$ produces locally as-rigid-as-possible (ARAP) motions to prevent over-fitting to the 3D scan input, which is formulated as:
\begin{align}
    E_{\text{reg}}^{\text{S}}(\mathcal{G}) = \sum_{k \in \mathcal{G}} \sum_{n \in \mathcal{N}_k} w_{k,n}^\mathcal{N} \Vert (\mathbf{g}_k - \mathbf{g}_n) - \mathbf{R}_k\left( \hat{\mathbf{g}}_k - \hat{\mathbf{g}}_n \right) \Vert ^ 2 _ 2,
    \label{eq4}
\end{align}
where $\mathcal{N}_k \in \mathcal{G}$ is the 1-ring neighborhood of the $k$-th ED node and $w_{k,n}^\mathbf{N}$ denotes the KNN weight between the $k$-th and $n$-th nodes.
For each ICP iteration, the resulting optimization problem in  Eq.(\ref{eq3}) is solved effectively using the Conjugate Gradient method.
Let $\mathbf{M}_S$ denote the vertices of the deformed template after the silhouette-based optimization.

\paragraph{Point cloud based deformation.}
After the above silhouette based alignment, we re-sample a finer ED graph to model the fine-detailed geometry information in the input scan.
For clarity and simplification, we reuse $\mathcal{G}$ to represent the ED motion from the previous results $\mathbf{M}_S$ to the input 3D mesh.
Then, the full energy function for current fine-detailed alignment is formulated as:
\begin{align}
    E_{\text{D}}(\mathcal{G}) = E_{\text{data}}^{\text{D}}(\mathcal{G}) + \lambda_{\text{reg}}^D E_{\text{reg}}^{\text{D}}(\mathcal{G}).
    \label{eq5}
\end{align}
Here, the data term $E_{\text{data}}^{\text{D}}(\mathcal{G})$ measures the fitting from $\mathbf{M}_S$ to the input mesh, which is formulated as the sum of point-to-point and point-to-plane distances:
\begin{align}
    \begin{split}
        E_{\text{data}}^{\text{D}}(\mathcal{G}) = & \lambda_{\text{point}}^{\text{D}}\sum_{i\in\mathbf{M}}\|\mathbf{v}_i(\mathcal{G}) - \mathbf{v}_i^c\|^2 + \\
        &\lambda_{\text{plane}}^{\text{D}}\sum_{i\in\mathbf{M}}(\mathbf{n}_{i}^\mathrm{T}(\mathcal{G})
        \cdot(\mathbf{v}_i(\mathcal{G})-\mathbf{v}_i^{c}))
    \end{split}
    \label{eq_pp_ppl}
\end{align}
where $\lambda_{\text{point}}^{\text{D}}$ and $\lambda_{\text{plane}}^{\text{D}}$ are the weights to balance two kinds of distances;
$\mathbf{n}_{i}(\mathcal{G})$ represents the normal of the deformed vertex $\mathbf{v}_{i}(\mathcal{G})$.
Note that for each $\mathbf{v}_{i}(\mathcal{G})$, its corresponding point $\mathbf{v}_{i}^c$ in the scan is found via the same look-up method in an ICP manner.

The regularity term $E_{\text{reg}}^{\text{D}}(\mathcal{G})$ here shares the same formulated as the one in Eq.(\ref{eq4}) and the full energy is solved using the same conjugate gradient solver.
After this point cloud based alignment on a finer scale, the vertices of the deformed template are denoted as $\mathbf{M}_D$.

\paragraph{Per-vertex deformation.}
Finally, we  refine the deformation from ED graph-based non-rigid result $\mathbf{M}_D$ to the input 3D mesh via per-vertex optimization, so as to improve the alignment accuracy, especially for those local regions with fine details like clothing wrinkle and boundary, which is formulated as follows:
\begin{align}
    E_{\text{V}}(\mathbf{M}) = & E_{\text{data}}^{\text{V}}(\mathbf{M}) + \lambda_{\text{reg}}^{\text{V}} E_{\text{reg}}^{\text{V}}(\mathbf{M}).
    \label{eq6}
\end{align}
Here, similar to Eq.(\ref{eq_pp_ppl}), the data term $    E_{\text{data}}^{\text{V}}$ further measures the per-vertex fitting by minimizing the both the point-to-point and point-to-plane distances:
\begin{align}
    \begin{split}
        E_{\text{data}}^{\text{V}}(\mathbf{M}) = &\lambda_{\text{point}}^{\text{V}}\sum_{i\in\mathbf{M}}\|\mathbf{v}_i - \mathbf{v}_i^c\|^2 + \\
        &\lambda_{\text{plane}}^{\text{V}}\sum_{i\in\mathbf{M}}(\mathbf{n}_{i}^\mathrm{T}\cdot
        (\mathbf{v}_i-\mathbf{v}_i^{c})).
    \end{split}
    \label{eq7}
\end{align}
We utilize the same regularity term $E_{\text{reg}}^{\text{V}}$ from \citet{xu2018monoperfcap} to prevent over-fitting to the 3D input scan.
let $\mathbf{M}_V$ denote the final optimized vertices of the human template.
Fig.~\ref{fig:3_stage} shows the intermediate alignment results of all these stages, which demonstrates the effectiveness of our multi-stage alignment scheme.
% Finally, we obtain a clothed deformed model, namely CDM.
%
After the multi-stage alignment, we obtain a deformed human template that is not only fitted to the captured 3D human scan but also owns the global consistent topology.
%We can calculate the tightness if given its the UnClothed Body Model, namely UCBM, the same deformed from CA-SMPL. 
%

\subsection{Geometry Image Representation}\label{GeoImgRep}
% The Following part is to establish the mapping from our template to a geometry image (UV map).
% \chen{Though it's possible to predict the tightness field directly on the 3D mesh using techniques like GCN~\citep{Litany_2018_CVPR,ZhouWLCYSLS20}, \minor{it consumes much more computational resources and is generally harder to train.}
% it consumes much more computational resources and is generally harder to train.
% Instead,
To benefit the tightness prediction with the effective image-to-image translation network, we map the clothed 3D human body mesh with the consistent topology into a regular 2D UV image, which has been proved to be effective in previous works, like \citet{Zorah2018ECCV, xu2019denserac} and \citet{alp2018densepose}.
\begin{figure}
    \centering
    \includegraphics[width= \linewidth]{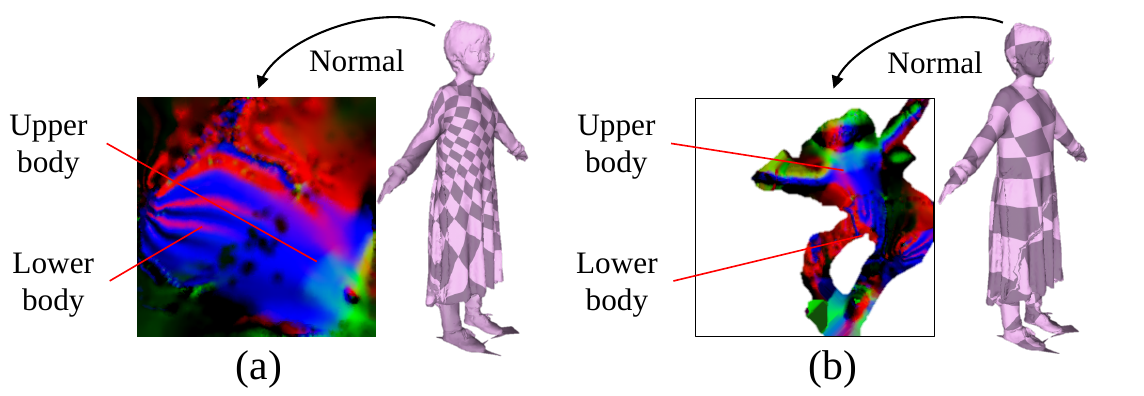}
    \caption{The comparison of two feature map with different mapping methods. (a) The normal map using geometry image~\citep{gu2002geometry}. (b) The normal map using OptCuts algorithm~\citep{OptCuts}. }
    \label{gi_feature}
\end{figure}

% XXXX? what pieces, don't understand.xxxx  GCN need to cut mesh int pieces
%Considering to maintain the clothing local continuity rather than cut mesh into pieces with GCN, we choose to map 3D body mesh to UV map, for generating feature map fed in our learning part (in Sec. \ref{TightNet}). It is an effective method to help learning-based methods extracting 3D feature, verified in \citet{Zorah2018ECCV, xu2019denserac, alp2018densepose}. However, many learning-based methods still combine several separate UV pieces. These pieces affect the continuity around the boundary of feature/UV maps. Inside of these methods, we choose geometry image~\citep{gu2002geometry} to generate our UV map.

There are many methods to generate a 2D mapping from a 3D mesh.
We choose two representative methods for comparison.
\label{R4Q1_3}
One is the mapping approach of the geometry image~\citep{gu2002geometry} with gapless filling but relative large distortions denoted as $M_{GI}(\cdot)$.
The other method is OptCuts~\citep{OptCuts}, denoted as $M_{Opt}(\cdot)$, which automatically seeks the best seam for cutting and generates an image with lower distortion but contains gap area.
Fig.~\ref{gi_feature} illustrates the mapping results for both methods, and we utilize the geometry image~\citep{gu2002geometry} to achieve a more smooth feature representation.

A quantitative comparison of these two methods for our full pipeline is provided in Sec.~\ref{TightNet_Evaluation}.
To generate consistent 2D feature embeddings, we map the positions, normals, and RGB colors of each vertex into its 2D map using the mapping approach~\citep{gu2002geometry}.
Linear interpolation is further conducted to fill the hybrid 2D clothed geometry image, which is denoted as clothed-GI.

%%%%%%%%%%%%%%%%%%%%%%%%%%%%%%%%%%%%%%%%%%%%%%%%%%%%%%%%%%%%%%%%%%%%%%%%%%%%%%%%%%%%%%%%%%%%%%%%%%%%%%%%%%%%%%%%%%%%%%%%
\section{Tightness Prediction}
Previous human reconstruction methods, including methods based on scanned depth map(s)\citep{new2011kinect, newcombe2015CVPR, collet2015high, Dou2016TOG}, and silhouette(s)\citep{baker2005shape, cheung2003shape, cheung2003visual, corazza2006markerless, mikhnevich2011shape, xu2018monoperfcap}, represent the human body as a single layer.
Recently, \citet{neophytou2014layered, pons2017clothcap, zhang2017detailed, yu2018doublefusion, yu2019simulcap} proposed the idea of multi-layer body shape recovery.
We extend this idea and define a novel clothing tightness formulation, which describes the relationship between the underlying human body shape and the various garment layers (Sec. \ref{tightnessMeasure}).
Subsequently, we propose a conditional GAN to predict the clothing tightness map in a data-driven manner, based on the 2D hybrid clothed-GI input and our novel tightness formulation.
(Sec.\ref{TightNet})
We also introduce an effective optimization scheme to reconstruct both the inner human shape and the multi-layer dressed garments accurately from the predicted tightness map in the geometry image domain (Sec. \ref{ShapRecovery}).

%
% Based on the hybrid feature map from the above clothed-GI, we propose to predict the corresponding 2D clothing tightness map in a data-driven manner, which utilizes a novel tightness formulation (see Sec.\ref{tightnessMeasure}) and an effective learning framework based on conditional GAN, named TightNet (see Sec.\ref{TightNet}).
\subsection{Tightness measurement}\label{tightnessMeasure}
\begin{figure}[t!]
    \centering
    \includegraphics[width= \linewidth]{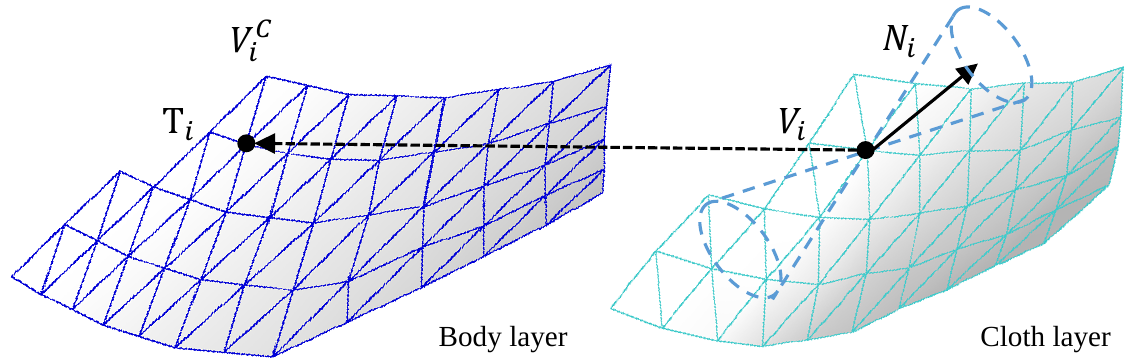}
    \caption{Illustration of tightness $\mathbf{T}_i$ of vertex $i$ on a mesh, which is the black dotted directed vector pointing from cloth to body.}
    %\textcolor{blue}{[Ma: Be more clear, like $T_i$ is the dotted directed vector...]}
    \label{fig_tightnessmeasure}
\end{figure}

% Straight-forward formulation
\chen{Different to previous human modeling methods~\citep{alldieck2018video,alldieck19cvpr} which model the displacements from the body template SMPL to the outer garment surface, we introduce the clothing tightness field to measure the displacements from various garments to the personalized underlying human body model.}
Recall that our clothing tightness dataset (CTD) contains both the dressed human meshes under various clothing and poses with the ground-truth underlying body models.
To model the tightness field from the clothing layer to the real body layer using our CTD dataset, a straight-forward formulation is to align our enhanced SMPL model in Sec.\ref{TemplateModel} to both the dressed mesh and the corresponding inner body mesh simultaneously using our multi-stage alignment method in Sec.\ref{alignment}.
%
% For simplification, let CDM 
%
Then, the per-vertex tightness is formulated on top of these two non-rigid aligned human templates which share the same topology:
\begin{align}
    \mathbf{T}_i & = \mathbf{v}_i - \mathbf{v}_i^c.
    \label{eq9}
\end{align}
Here, $\mathbf{v}_i$ and $\mathbf{v}_i^c$ are the $i$-th vertex of the two templates aligned to the clothing layer and the body layer, respectively.
%
%Tightness indicates the displacement between the real body layer and cloth layer rather than the displacement between the reconstructed/aligned mesh and template mesh.
%其中表示该点$v_i$在对应不穿衣服的人体模型(unclothed mesh) 的对应点。$F_i\in \mathbb{R}^3$ 是一个从穿衣模型顶点指向不穿衣服模型顶点的向量，该向量的模表示对应两点间的欧拉距离。
%We define the tightness on mesh with the topology of our CA-SMPL. As we mentioned, the tightness is the gap between two layers (body/cloth), so we take our UCBM (aligned with an unclothed body mesh) and CDM (aligned with a clothed mesh) as an example.
%
Note that the direction of $\mathbf{T}_i$ indicates the corresponding pairs from clothing to the inner body shape, while its magnitude is the euclidean distance between the two corresponding vertices.
We can further define the tightness field, denoted as the matrix $\mathcal{T}$, on top of our enhanced SMPL model as follows:
\begin{align}
    %\begin{split}
    \mathbf{\mathcal{T}} & = \left\{ \mathbf{T} \in \mathbb{R}^{N_{\mathbf{M}} \times 3} \right\},
    %\end{split}
    \label{eq8}
\end{align}
where $N_{\mathbf{M}}$ is the number of the template vertices, same as Eq. \ref{eq0}.

% More complicated formulation
% since humans generally wear more casual clothing than compression or fit. The relation between body shape and its cloth layer also relies on many factors, eg., body pose, motion, clothing materials, thickness, etc.

%为了在真实的数据中计算tightness, 我们首先利用已知的Cloth label数据，将MVS　模型上的点、面分为衣服部分和皮肤部分
%再从裸体模型上的每一个顶点v_t出发，以这个顶点的法向n_t和其反方向n_t*为中心方向，向周围15度方向发射射线，如果射到MVS模型上属于衣服的平面f_c ，
%则建立v_t与f_c上顶点v_c 的映射关系。在迭代完成后，对每一个衣服上的顶点v_c我们利用guassian kernel 对所有于它建立映射关系的点进行去噪，保留其中相对距离较小部分
%之后，对于每一个没有和任何皮肤上的点建立映射关系的衣服上的点v_c，我们迭代衣服上的点v_t，计算v_c 与v_t之间的距离，并选择距离v_c距离最小的点作为它的对应点。
%最后，我们计算与v_c对应的v_t的均值v_t avg,tightness 即等于v_c 与 v_t avg之间的offset

%
However, this straightforward formulation above fails to model the exact correspondences between the clothing layer and the body layer, because the two non-rigid alignments from the template to the dressed mesh and the inner body model are performed totally independently, and the one-to-one correspondences are fragile to the alignment error.
To this end, we formulate the per-vertex tightness as the one-to-many correspondences between the human template aligned to the clothing layer and the ground-truth body model directly, jointly considering the direction and distance information of the clothing layer.
For a vertex $\mathbf{v}_i$ on the aligned human template of the dressed mesh, we calculate its approximated tightness $\hat{\mathbf{T}_i}$ as follows:
%
%In practice, we use the CDM (cloth layer) to enable calculating the ground truth tightness between its and the carved shape mesh (not a UCBM from our CA-SMPL) directly. 
%
\begin{align}
    \hat{\mathbf{T_i}} = \frac{\sum_{\mathbf{v}_r^c \in \mathcal{N}_1^c}K_G(\mathbf{v}_i -\mathbf{v}_r^c)+\sum_{\mathbf{v}_d^c \in \mathcal{N}_2^c}K_G(\mathbf{v}_i-\mathbf{v}_s^c) }{\left|\left| \mathcal{N}_1^c\right|\right|+\left|\left| \mathcal{N}_2^c\right|\right|}.
    \label{eq10}
\end{align}
Here, the one-to-many correspondence set \(\mathcal{N}_1^c\) denotes the local vertices set of the ground-truth human body shape found via a ray-tracing operation along the normal direction of $\mathbf{v}_i$ within a double-cone with an aperture of 30 degrees, while \(\mathcal{N}_2^c\) is the set of the 20 closest vertices of $\mathbf{v}_i$ in the ground-truth body mesh in terms of Euclidean distance, as shown in Fig.\ref{fig_tightnessmeasure}.
% (R1) Using surface normal on scan is not reliable to find corresponding body region as surface normal can be very noisy due to wrinkles
% 解释: 所以我们不仅仅使用了normal采用 顶点欧拉距离结合法向的加权，法向的权重较小
%       如果仅仅采用欧拉距离，不利用法向，在肩部 膝盖 拐角处 最近点容易收敛到单一位置？ xxx
%       为什么一定要用法向？？？ 再思考下
Note that $K_G(\cdot)$ is the Gaussian weighting function based on the angle between two vertex normals to enable smooth tightness field estimation.

\chen{After the above tightness field estimation from the dressed template to the ground-truth body model, we further apply the same strategy but change the target and source. We calculate this per-vertices tightness from the template for more reliable correspondence and combine such bi-directional estimations to obtain the ground-truth 3D clothing tightness field with the same topology of the enhanced SMPL template.}
\chen{Specifically, we utilize a linear weighting for two tracking results, \minor{which is usually 0.8 for the body template to the dressed template and 0.2 for the other.}}
Then, by using the same mapping operation in Sec.\ref{GeoImgRep}, we generate the tightness map in the geometry image domain so as to enable end-to-end learning of the clothing tightness field and implicitly to model the influence of various clothing categories, looseness, and human postures.

\subsection{TightNet architecture}\label{TightNet}
\chen{Based on the tightness map above and the hybrid 2D feature map from Sec.~\ref{GeoImgRep} in the global geometry image domain, we thus propose to train a pix2pix-style~\citep{pix2pix} convolutional neural network, which is the most effective image-to-image translation network structure verified in many previous works \citep{lazova2019360, pumarola20193dpeople, alldieck2019tex2shape}.} We denoted this net as TightNet, to infer the clothing tightness map and garment masks in an end-to-end manner.
In the following, we provide more details about the input/output pair, the used network architectures, losses, and training schemes.

% Input & output
The input to our TightNet is the hybrid feature embedding in the clothed-GI from the raw 3D scan, including the vertex positions, normals, and RGB colors, while the output consists of the predicted tightness map as well as the masks for both the upper and lower garments, so as to enable further multi-layer garment reconstruction.
\label{explain_mask}
% R5Q2 R5Q3:Prediction and supervision of clothing masks is not explained.
% A:进一步解释衣服的mask是怎么样监督和预测的 放到method这里说明 实验部分说明? mask是怎么算出来
\pang{Note that for those clothing categories, we set two main categories, upper garment and lower garment, including shirt, coat, jacket, and dress for upper garment, pant and skirt for the lower garment. We model the upper garment and the lower garment but take the whole dress (not including the skirt) as one upper garment.
For the prediction of garment masks, we utilize the same TightNet to predict tightness map with the five channels of predicted results (three channels for tightness map, two channels for the mask of upper/lower garments), and supervise the mask with the provided segmented garment in our dataset. We use the L1-loss for the garment mask training, which is the same as the tightness map supervision.}
Thanks to our unique 2D mapping scheme based on geometry image~\citep{gu2002geometry}, both the input and output share the same semantic 2D structure, so as to implicitly handle the huge variations for clothing categories, garment looseness, and human postures.

% Our architecture
The network in our TightNet is a conditional Generative Adversarial Network (Pixel2Pixel)~\citep{pix2pix}, \pang{which learns a mapping from the input hybrid feature map to our tightness map and mask map.
% note:Pix2pix does not take random noise vector as input unlike DC-GAN
%
More specifically, the generator is U-Net~\citep{ronneberger2015u} encoder-decoder structure with skip connections between convolution-ReLU-batch norm down- and up-sampling layers, which can share information between the input and output. The input is the nine channels of hybrid feature map with 224$\times$224 resolution including every three channels for vertex positions, normals, and RGB colors, while the output is the five channels of predicted results with the same resolution including three channels for tightness map, two channels for upper/lower clothing mask maps.}

\pang{\label{R1Q6_16}In our discriminator, we utilized PatchGAN~\citep{pix2pix} discriminator. However, we take this architecture to predict the full-body tightness field 
%learn the mapping from input hybrid feature map to tightness/mask map, 
rather than image style transfer. Unlike the original PatchGAN~\citep{pix2pix} to restrict their attention to small local patches, \minor{we take the full feature map as the random patches and further normalize these patches in our GAN discriminator.}}
% xxxxxxxxxxxxxx
% (old version) Since the high-frequency structure is not important in our task, we further normalize the full image GAN discriminator with a pyramid structure.
%
% our loss and training details
We train the TightNet with the well-established L1-loss instead of L2-loss for fewer blurring artifacts.
Benefiting from our tightness predictor, we can extract the hidden information between different clothing appearances and the tightness, while the input positions and normals also help our predictor to consider the effect of the current human pose.

%\begin{equation}
%    \begin{aligned} \mathcal{L}_{G A N}(G, D)=& \mathbb{E}_{x, y}[\log D(x, y)]+\\ & \mathbb{E}_{x, z}[\log (1-D(x, G(x, z))]\end{aligned}
%\end{equation}\noindent where $G$ minimizes $\mathcal{L}_{ G A N}$ while the adversarial $D$ maximizes it. 
%\begin{equation}
%    \mathcal{L}_{L 1}(G)=\mathbb{E}_{x, y, z}\left[\|y-G(x, z)\|_{1}\right]
%\end{equation}
%begin{equation}
%    G^{*}=\arg \min _{G} \max _{D} \mathcal{L}_{ G A N}(G, D)+\lambda \mathcal{L}_{L 1}(G).
%\end{equation}

\subsection{Shape recovery under clothing}\label{ShapRecovery}
% \textcolor{blue}{[Ma: It is not a sentence]}.
\chen{To fine-tune the predicted body from TightNet, especially fixing the noise around the local regions like the oxter and crotch, we utilize both the predicted tightness field and the prior of the warped template.}
Thus, based on the clothing tightness and mask prediction above, we propose an effective optimization scheme to reconstruct both the inner human shape and the multi-layer dressed garments accurately.

\paragraph{Shape recovery.}
Recall that our clothing tightness field indicates the displacements from the garment layers to the inner human body layer.
To cover the body shape from the tightness prediction, we first utilize the inverse function of the mapping $M_{GI}^{-1}(\cdot)$ in Sec.\ref{GeoImgRep} to generate the per-vertex
tightness field $\hat{\mathcal{T}}$ on the final aligned template mesh $\mathbf{M}_V$, where $\mathbf{\hat{\mathcal{T}}} = \left\{ \hat{\mathbf{T}} \in \mathbb{R}^{N_{\mathbf{M}} \times 3} \right\}$.
Then, a straight-forward solution to obtain the inner body shape $\mathbf{M}$ based on our tightness field formulation is as follows:
\begin{align}
    %\begin{split}
    \mathbf{M} & = \mathbf{M}_V + \hat{\mathbf{T}}.
    %\end{split}
    \label{bodyDirect}
\end{align}
However, such solution above suffers from the tightness field estimation error, especially for those local regions around oxter and crotch under unusual human poses, leading to visually unpleasant body shape recovery.
To this end, we propose a simple and effective optimization scheme to estimate a smoother body shape by solving the following least-squares energy function:
\begin{align}
    \begin{split}
        E_{\text{body}}(\mathbf{M}) = & \lambda_{fit}(\mathbf{M}_V+\mathbf{T}-\mathbf{M}) + \\& \lambda_{smooth}(\mathbf{M}-K_G(\mathbf{M})) +\\& \lambda_{reg}(\mathbf{M}-\mathbf{M}_{warp}).
    \end{split}
    \label{bodyShapeOptimization}
\end{align}
Here the first data term utilizes our tightness field formulation similar to Eq.(~\ref{bodyDirect}), while the second term enables smooth body shape estimation via the same Gaussian kernel $K_G(\cdot)$ defined in Eq.(\ref{eq10}).
In the final regular term, the warped vertex matrix $\mathbf{M}_{warp}$ denotes the warped body template after the first ICP iteration of the first stage optimization in our multi-stage alignment in Sec.~\ref{alignment}.
Such regular term forces the optimized body shape to be closed to the utilized human template to penalize unnatural body shapes.
All the parameters for these three terms are empirically set to be is 1, 0.1, and 0.05, respectively.
Finally, by solving the least-squares problem in Eq.(\ref{bodyShapeOptimization}), we reconstruct an accurate and visually pleasant body shape of the input 3D human scan.

%\begin{align}
%    %\begin{split}
%    \mathbf{\hat{\mathcal{T}}} & = \left\{ \hat{\mathbf{T}} \in \mathbb{R}^{N_{\mathbf{M}} \times 3} \right\}.
%    %\end{split}
%\end{align}

\paragraph{Clothing segmentation.}
Besides the body shape recovery above, we utilize the output of multiple garment masks from our TightNet to automatically segment clothing from the human body on 3D meshes so as to enable further cloth retargeting or animation applications.
Since the output masks are not accurate enough to segment the clothing directly in the 3D space, we utilize the following Markov Random Fields in \citet{pons2017clothcap} to solve the per-vertex clothing label $v_i\in\mathbf{v}$ for each vertex in our final aligned template mesh $\mathbf{M}_V$:
\begin{align}
    \begin{split}
        E_{\text{cloth}}(\mathbf{v}) = \sum_{i \in \mathcal{T}} \varphi_{i}\left(v_{i}\right)+\sum_{(i, j) \in \mathcal{T}} \psi_{i j}\left(v_{i}, v_{j}\right).
    \end{split}
\end{align}
To enable fully automatic segmentation, we replace the manually defined garment prior of the original optimization in \citet{pons2017clothcap} with our predicted garment masks from TightNet.
Please kindly refer to \citet{pons2017clothcap} for more details about how to solve the energy function above.

%%%%%%%%%%%%%%%%%%%%%%%%%%%%%%%%%%%%%%%%%%%%%%%%%%%%%%%%%%%%%%%%%%%%%%%%%%%%%%%%%%%%%%%%%%%%%%%%%%%%%%%%%%%%%%%%%%%%%%%%
\section{Experimental Results}
In this section, we evaluate our method on a variety of challenging scenarios.
We first report the implementation of the details of our whole method and two utilized datasets, followed by the evaluation of our main technical contributions. We also include both qualitative and quantitative comparisons with previous state-of-the-art methods.
The applications and limitations regarding our approach are provided in the last two subsections.

%First of all, we present a 3D human and clothing dataset, called the Clothing Tightness Dataset (CTD).  %Our experiment is conducted on our CTD, and the Bodies Under Flowing Fashion (BUFF) dataset proposed in~\citet{zhang2017detailed}. 

\begin{figure}[t!]
    \centering
    \includegraphics[width=0.95\linewidth]{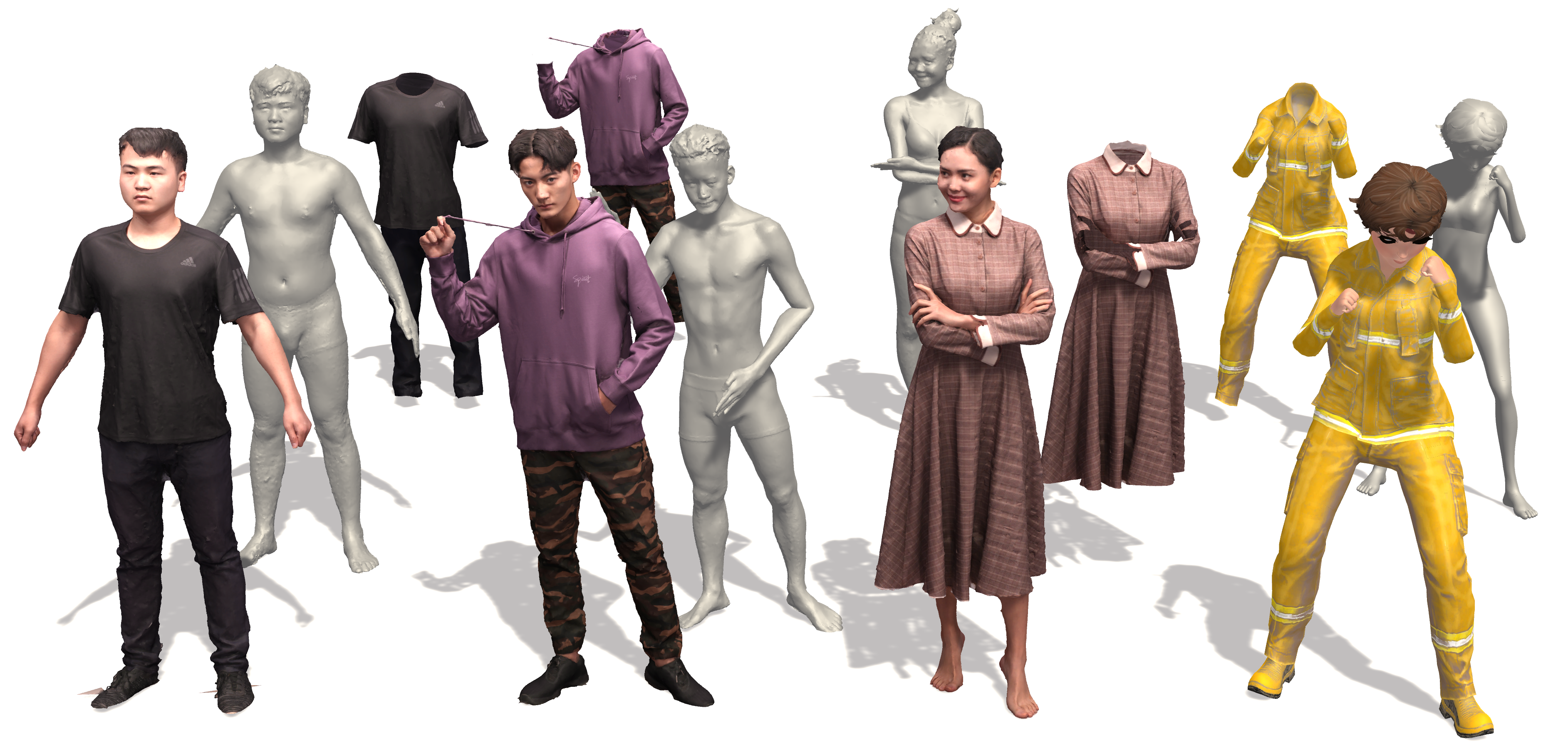}
    \caption{Sample data from our dataset: three real human subjects with scanned body shape meshes, the segmented clothes, and one synthetic model (rightmost) for pre-training. }
    \label{fig:dataset}
\end{figure}

\begin{figure*}[t]
    \centering
    \includegraphics[width=0.98\linewidth]{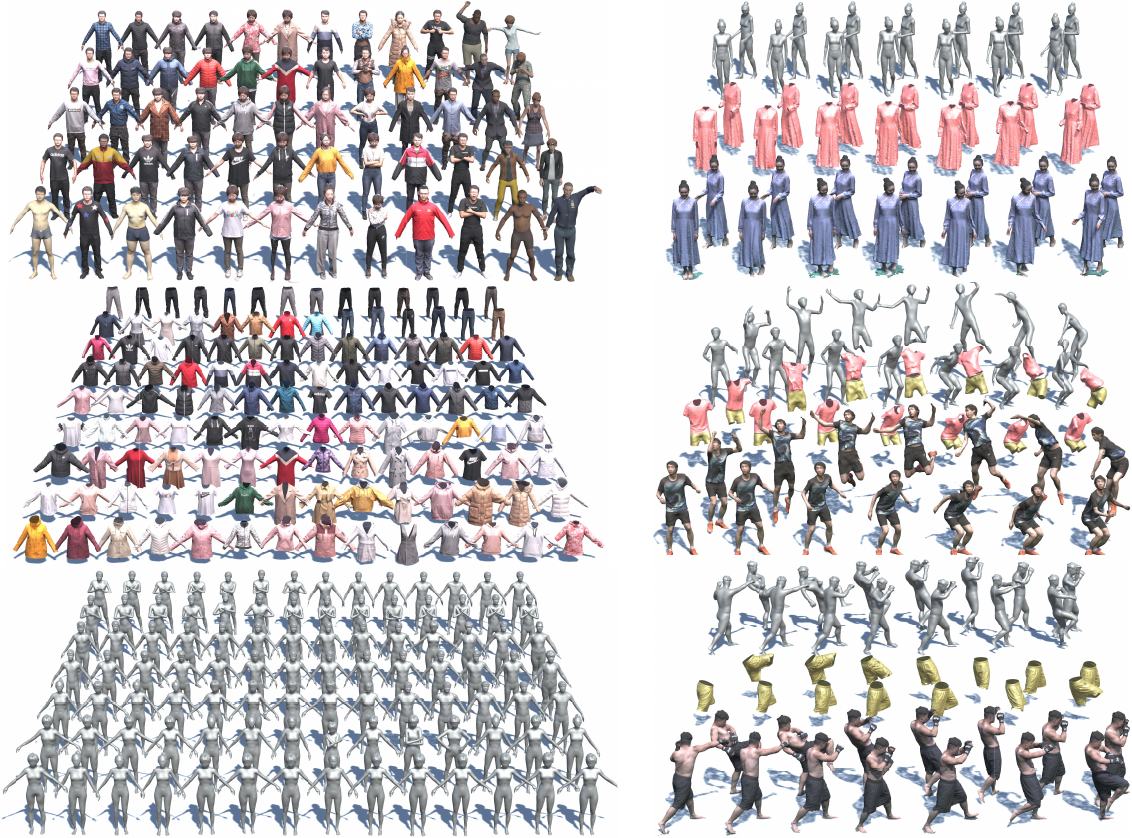}
    \centering
    \caption{The gallery of our Clothing Tightness Dataset (CTD). The first column from top to bottom: 1) Sampling of various clothed human, including synthetic models (rightmost) and two body shape scans. 2) Various segmented clothes with 'A' pose only. 3) Carven body shapes with 'A' pose only. The second column is three typical dynamic sequences in our dataset including clothed human, segmented clothes, and carven body shapes.}
    \label{dataset_gallery}
\end{figure*}

\begin{figure*}[t!]
    \centering
    \includegraphics[width=0.98\linewidth]{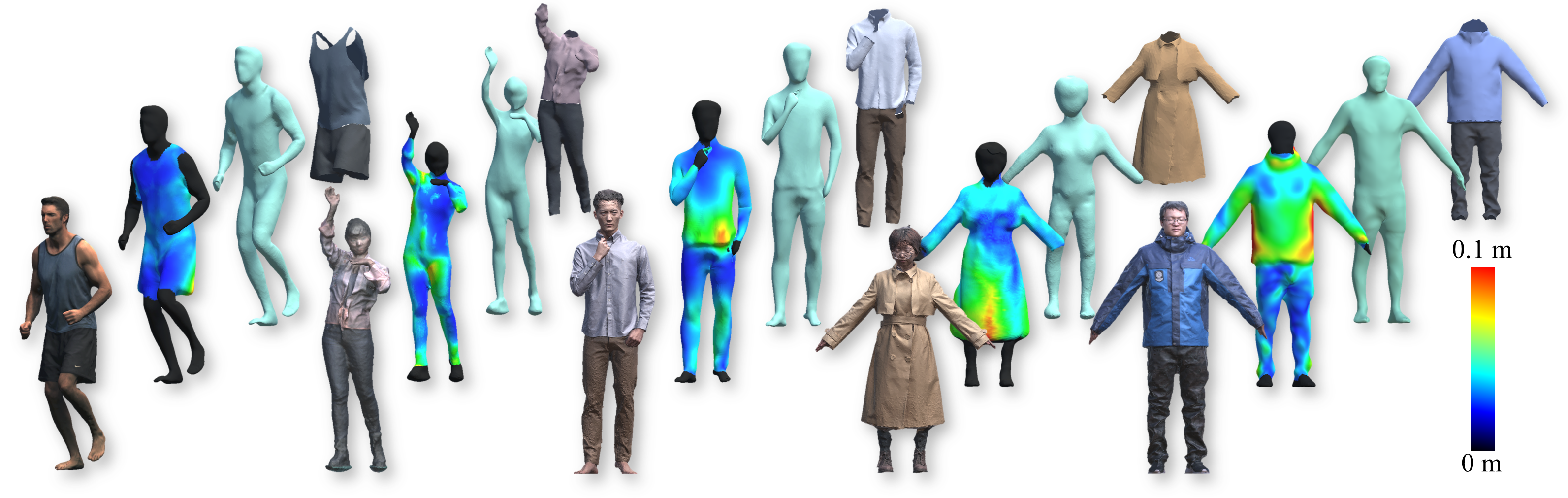}
    \caption{\chen{The gallery of our results. From down to top, the captured meshes, predicted tightness field, recovered body shapes and segmented clothes.}}
    \label{fig:gallery}
\end{figure*}

\paragraph{Implementation details.}
We run our experiments on a PC with an NVIDIA GeForce GTX 1080Ti GPU, a 4.2 GHz Intel Core i7-7700K CPU, and 16 GB RAM.
Our optimized code takes about 12 s per 3D human scan, which divides to 5-8 s for the multi-stage human template alignment (about 1.5, 2, and 3 s for each stage, respectively), 0.5 s for the tightness map prediction, and 0.5 s for the shape recovery from tightness field.
All the energy functions of the multi-stage alignment are solved via a GPU-based Conjugate Gradient solver.
%
% For the computation time in our algorithm, we implement an efficient Gauss-Newton solver based on CUDA for our optimizations. 
%
In all experiments, we use the following empirically determined parameters:
$N_{\mathbf{G}^S} = 1407$,
$\lambda_{reg}^S = 10$;
$N_{\mathbf{G}^D} = 2103$,
$\lambda_{reg}^V = 7$
$\lambda_{point}^D = 0.5$,
$\lambda_{plane}^D = 1.5$;
$\lambda_{reg}^V = 1$,
$\lambda_{point}^V = 1$ and
$\lambda_{plane}^V = 1.5$.
For the tightness prediction, we set the resolution of clothed-GI as 224$\times$224. Note that the clothing tightness predictor is pre-trained on our dataset, which takes about 3 hours, and the training on CTD needs 4 hours.

\subsection{Dataset}\label{dataset}
For a thorough evaluation of our method, we utilize both the most popular public dataset and a much larger captured dataset for the task of reconstructing both the human body shape and garments.

\paragraph{Bodies Under Flowing Fashion Dataset.}
The BUFF dataset ~\citep{zhang2017detailed} is the most popular public dataset for body shape estimation, which contains three males and three females models wearing two types of clothing (t-shirt/long pants and a soccer outfit).
It provides a dynamic sequence for each subject but only with the per-vertex color rather than extra high-quality RGB textures.
BUFF also contains the body shapes under the general T pose without garments as the ground truth.
Since the data size of BUFF is far from enough to train our tightness prediction network, we only utilize the sampling frames from their scans as input and predict the tightness with the pre-trained model using our dataset.

% a new 3D dataset that consists of a large variety of clothing including T and long shirt, short/long/down coat, hooded jacket, pants, skirt/dress, etc., and the corresponding 3D human shapes. 

\paragraph{Clothing Tightness Dataset (CTD).}
To model clothing tightness in a data-driven manner, we propose a new benchmark dataset, which contains 880 dressed human meshes with both the body geometry and segmented individual pieces of garments.
Among them, 228 meshes are statically captured, and 652 are from dynamic 3D human sequences (13 sequences in total).
\chen{We have captured 18 subjects, 9 males and 9 females, 10 of them are with the canonical "A" or "T" poses and 8 subjects are under dynamic daily actions, including boxing, dancing, playing badminton, keep-fit exercise, and so on.}
For garment modeling, our CTD contains 228 different garments for each static caption, including T/long shirt, short/long/down coat, hooded jacket, pants, and skirt/dress, ranging from ultra-tight to puffy.
For each dynamic sequence, we capture 400\textasciitilde500 frames under 30 fps and evenly sample 40\textasciitilde50 frames for our dataset.
Note that most 3D meshes are reconstructed via a dome system equipped with 80 RGB cameras using the MVS approach~\citep{schonberger2016pixelwise}, with about 50,000 vertices, 100,000 facets, and a 4K texture, while few 3D meshes are reconstructed via the DynamicFusion approach~\citep{newcombe2015CVPR} with very similar quality.
The corresponding ground truth naked human shapes are obtained via the same system, and then 5 artists further manually segment each piece of clothing and carve the body shape out from the raw mesh.
We then generate the ground truth tightness field using the novel formulation in Sec.\ref{tightnessMeasure}.
Fig.~\ref{fig:dataset} illustrates the high-quality examples from our dataset, while Fig.~\ref{dataset_gallery} further provides the gallery of our whole dataset.
We will make our dataset publicly available.
\chen{To train our tightness prediction network in Sec.~\ref{TightNet}, we split the data into 80\% vs. 20\% with considering the identities. With more dynamic frames for the training set to provide more training instances, we keep half of the identities that do not appear in the training set.}
We also generate 800 clothed human meshes with synthetic avatars using Adobe Fuse CC for the pre-training of our network.
Fig.~\ref{fig:gallery} demonstrated the multi-layer results of our approach, where both the human body shape and the garments under various clothing tightness and human postures are faithful reconstructions.

\subsection{Evaluation}
In this section, we evaluate our individual technique components, i.e., the human template alignment, the clothing tightness prediction, as well as the shape recovery from the tightness map in the following contents, respectively.

% A.1 enhanced-SMPL vs SMPL  alignment对比
\begin{figure}[t]
    \centering
    \includegraphics[width=\linewidth]{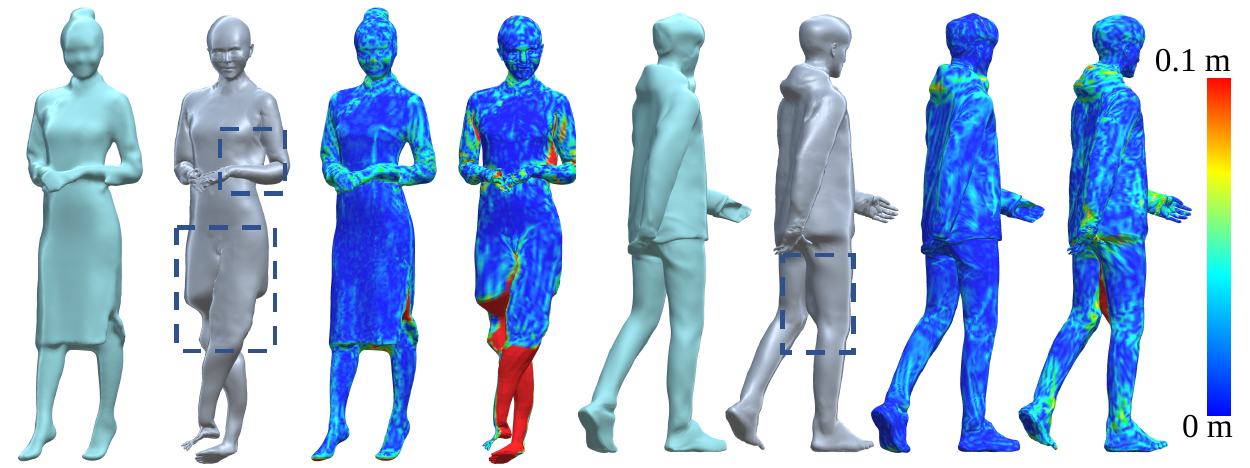}
    \caption{\chen{The comparison between our enhanced clothing-adapted SMPL (CA-SMPL) and the high resolution SMPL \citep{loper2015smpl} (27554 vertices) for the clothing alignment. The green meshes are the results of our alignment algorithm using our enhanced SMPL, while the gray meshes are the results of the same algorithm using SMPL. The alignment error is color-coded from blue to red.}}
    \label{Comparison_SMPL}
\end{figure}

\paragraph{Alignment evaluation.}
We first evaluate the effectiveness of our clothing-adaptive human template (CA-SMPL) in Sec.~\ref{TemplateModel} by comparing it to the original SMPL model~\citep{loper2015smpl} using the same multi-stage alignment algorithm.
As shown in Fig.\ref{Comparison_SMPL}, the original SMPL suffers from severe alignment error, especially for those local regions like crotch and neck due to the limited generation ability to handle clothing variations.
\chen{In contrast, our enhanced template is more robust to both the clothing variations in our dataset, leading to the improvement in the clothing alignment accuracy.}
%
%and template model should be more robust for these input meshes with missing geometry detail. 
%

% A.2 three-Stage 的优化每个stage的作用
% A.3 甚至和ICCV review 挑刺和XXX太像了的那种baseline的alignment对比
\label{3_Stage_rs}
\begin{figure}[t!]
    \centering
    \includegraphics[width=\linewidth]{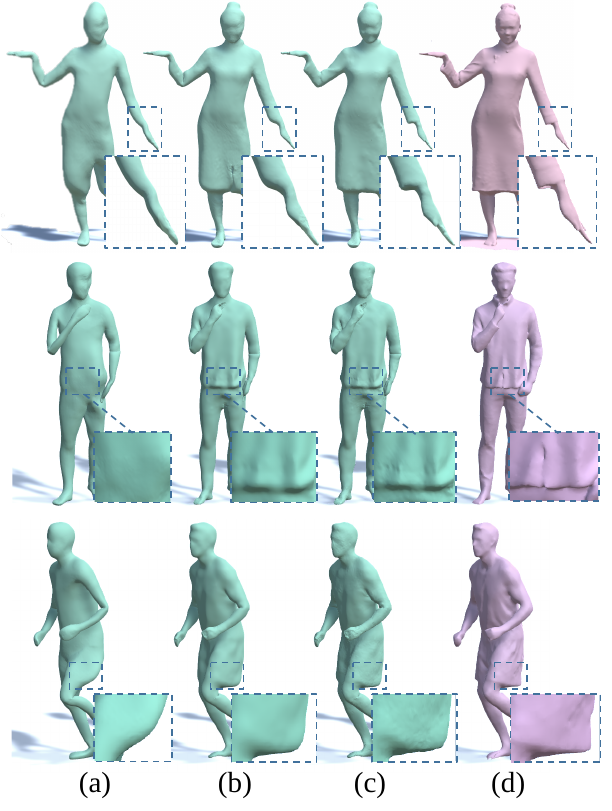}
    \caption{The qualitative evaluation of our multi-stage alignment method. (a) The results after silhouette based alignment. (b) The results after point cloud based alignment. (c) The results after per-vertex alignment. (d) The captured meshes (Target meshes).}
    \label{3_stage_more}
\end{figure}

\begin{figure}[t!]
    \centering
    \includegraphics[width=\linewidth]{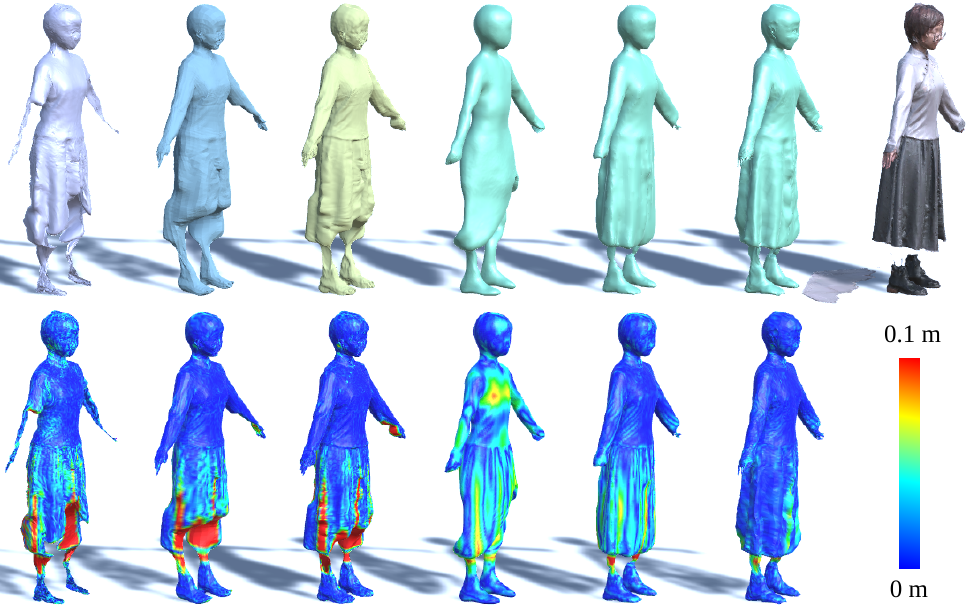}
    \caption{\minor{The qualitative comparison of our multi-stage alignment method on per-vertex error. Non-rigid \citep{tong2012scanning} is a baseline to directly align a 3D mesh with the captured mesh. From left to right, aligned mesh using Non-rigid \citep{tong2012scanning} (gray), SMPL based non-rigid method \citep{lazova2019360} (blue), implicit function based registration IP-Net \citep{bhatnagar2020ipnet} (yellow), our aligned mesh with silhouette, our aligned mesh with point cloud, our final aligned mesh, and target clothed mesh. The second row is per-vertex alignment error colored from blue (good) to red (bad).}}
    \label{3_stage_non_rigid}
\end{figure}

\begin{table}[t!]
    \centering
    \begin{tabular}{l|c|c|c}
        \hline
        Method                                    & Mean$\downarrow$ & RMS$\downarrow$  & Error (mm)$\downarrow$ \\ \hline \hline
        Non-rigid\citep{tong2012scanning}         & 0.448\%          & 0.762\%          & 13.44                  \\ \hline
        \chen{\citet{lazova2019360}}                     & \chen{0.340\%}          & \chen{0.563\%}          & \chen{10.02}                  \\ \hline
        \chen{IP-Net\citep{bhatnagar2020ipnet}}          & \chen{0.324\%}          & \chen{0.528\%}          & \chen{9.72}                   \\ \hline
        Silhouette based\citep{xu2018monoperfcap} & 0.494\%          & 0.779\%          & 14.82                  \\ \hline
        Point cloud based                         & 0.286\%          & 0.585\%          & 8.58                   \\ \hline
        Ours                                      & \textbf{0.263\%} & \textbf{0.521\%} & \textbf{7.89}          \\ \hline
    \end{tabular}
    \caption{Comparison of alignment methods for clothed human mesh. \emph{Non-rigid}~\citep{tong2012scanning} is a baseline to directly align a 3D scanned mesh with the input point cloud. \chen{\citet{lazova2019360} is a SMPL based registration to align both shape/pose parameters before non-rigid deformation. IP-Net\citep{bhatnagar2020ipnet} is a implicit function based registration with SMPL model.} \emph{Silhouette based}~\citep{xu2018monoperfcap} is a baseline to align a 3D mesh from silhouette only, which is also our first stage. \emph{Point cloud based} is our second stage using the results of the first stage as an initial value. $\downarrow$ means the smaller is better. \emph{Mean} and \emph{Root-Mean-Square (RMS)} are the metrics of Hausdorff distance~\citep{cignoni1998metro} from the sampling of the targets, normalized with the bounding box diagonal of all clothed meshes, which is 3 in our setting. \emph{Error (mm)} represents the per-vertex error with millimeter, using the same Hausdorff distance.}
    \label{3Stage_table}
\end{table}
\begin{table}[t!]
    \centering
    % \color{Blue}
    \begin{tabular}{l|c|c|c|c}
        \hline
        Method                                    & Mean$\downarrow$ & RMS$\downarrow$  & \tabincell{c}{Error\\ (mm)}$\downarrow$ & \tabincell{c}{Cor. Error\\ (mm)}$\downarrow$ \\ \hline \hline
        Silhouette based & 0.316\%          & 0.426\%          & 9.48                   & 24.37                  \\ \hline
        Point cloud based                         & 0.083\%          & 0.186\%          & 2.49                   & 16.30                  \\ \hline
        Ours                                      & \textbf{0.081}\% & \textbf{0.157}\% & \textbf{2.43}          & \textbf{13.93}         \\ \hline
    \end{tabular}
    \caption{\chen{Evaluation of our multi-stage alignment method on the FAUST \citep{Bogo:CVPR:2014} dataset. \emph{Silhouette based}~\citep{xu2018monoperfcap} is a baseline to align a 3D mesh from silhouette only, which is also our first stage. \emph{Point cloud based} is our second stage using the results of the first stage as an initial value. $\downarrow$ means the smaller is better. \emph{Mean}, \emph{Root-Mean-Square (RMS)}, and \emph{Error (mm)} same as Tab. \ref{3Stage_table}, and \emph{Cor. Error (mm)} represents the avenge of the per-vertex error with ground truth of the  correspondence vertices in FAUST \citep{Bogo:CVPR:2014}.}}
    \label{3Stage_table_faust}
\end{table}

We further evaluate our multi-stage alignment in Sec.~\ref{alignment} by analyzing the influence of each stage, both qualitatively and quantitatively.
Let \emph{Silhouette based} and \emph{Point cloud based} denote the variations of our alignment method after the first and the second stages, respectively.
Besides, we further compare to the alignment baseline~\citep{tong2012scanning}, which directly aligns a 3D mesh with the input point cloud, denoted as \emph{Non-rigid}.
In Fig.\ref{3_stage_more} we present the qualitative results of each stage for various challenging inputs.
Note that our full scheme achieves superior alignment results and can even float the crack on skirt and match the clothing boundary around the wrist.
%the first stage (silhouette based) can only generally align clothing from the view of the virtual camera, but the limbs have already fitted. Although the clothing details in the second stage (point cloud based) are satisfactory, the three-stage (per-vertex) 
%
Furthermore, the qualitative and quantitative results in Fig.~\ref{3_stage_non_rigid} clearly demonstrate the effectiveness of each stage in our alignment scheme.
Meanwhile, without the good initial state provided by the silhouette based deformation, the baseline~\citep{tong2012scanning} can not converge to a good result.
%
%For more different clothing types, we show the result in Fig.\ref{3_stage_more}. Our method can adapt to various clothing types. 

For further quantitative evaluation on our dataset, we utilize Metro~\citep{cignoni1998metro}, which is based on Hausdorff distance for comparing the difference of two meshes, and calculate its normalized \emph{Mean} and \emph{Root-Mean-Square (RMS)} as the metrics with a normalized factor (3 in our setting).
We also calculate the per-vertex error as a relative quantitative metric, denoted as \emph{Error (mm)}.
Tab.\ref{3Stage_table} shows that our full pipeline consistently outperforms the other baseline variations in terms of all these quantitative metrics.
\chen{We also compare with two SMPL based registration approaches used in \citet{lazova2019360} and IP-Net\citep{bhatnagar2020ipnet}.
They are based on shape/pose parameters fitting and non-rigid deformation with displacement from original mesh or reconstructed mesh from implicit function.
However, we utilize not only the original 80 cameras but also the 30 synthetic cameras (see Sec. 4.3) to support the initial 3D poses, and the silhouette and point cloud also provide coarse-to-fine references.
Thus, our multi-stage alignment produces 2 to 3 millimeters improvement on accuracy.}
% \citet{lazova2019360} 参考Major说明SMPL与我们的数据并不适用    IPnet的implict function 确实有非常优秀的alignment的效果，甚至比 Non-rigid based approach 更为鲁棒，考虑到non-riggide需要非常好的good initialization close to the data
% 但是IPnet依然是基于SMPL+displacement 考虑到SMPL不适用于我们的数据集,所以我们的效果更好
%

\chen{We then evaluate each alignment stage's correspondences error on FAUST \citep{Bogo:CVPR:2014} quantitatively.
The FAUST \citep{Bogo:CVPR:2014} dataset provides the ground-truth correspondences although the models are unclothed.
As shown in Tab. \ref{3Stage_table_faust}, each stage of our method gets a 2 to 8 cm decrease for the correspondence error.}

This not only highlights the contribution of each alignment stage but also illustrates that our approach can robustly align the enhanced human template to the input 3D scan.

% As shown in Tab.\ref{3Stage_table}, compared to directly non-rigid alignment~\citep{tong2012scanning} method, we improve 5.55 mm on the mean error of alignment. This evaluation shows our alignment method can be more robust and accurate than the directly non-rigid alignment.

\paragraph{TightNet evaluation.}\label{TightNet_Evaluation}
Here, we evaluate our TightNet quantitatively by comparing with two variation baselines using L2 loss or OptCut~\citep{OptCuts} for 2D mapping, denoted as \emph{Baseline L2} and \emph{OptCuts L1}, respectively.
We utilize the L1 norm and the structural similarity (SSIM)~\citep{wang2004image} for predicting the perceived quality of images, with window size (11 in our setting) to avoid the unreasonable effectiveness~\citep{zhang2018unreasonable}.
To evaluate the garment mask prediction, we utilize the mask IoU with a threshold of 0.5 since the mask output of TightNet is 0 to 1 initially. 
\begin{table}[t]
    \centering
    \begin{tabular}{l|c|c|c}
        \hline
        Method      & SSIM$\uparrow$   & L1/L2$\downarrow$ & mask IoU$\uparrow$ \\ \hline \hline
        Baseline L2 & 62.27\%          & 0.281           & 90.17\%            \\ \hline
        %OptCuts with mask L1 & 64.82\%          & 0.1892          & 85.32\%          \\ \hline
        OptCuts L1  & 43.91\%          & 0.493          & 88.20\%            \\ \hline
        Ours  L1    & \textbf{67.24\%} & \textbf{0.222} & \textbf{93.89\%}   \\ \hline
    \end{tabular}
    \caption{Evaluation of our TightNet. $\uparrow$ means the larger is better, while $\downarrow$ means the smaller is better.}
    \label{GAN_table}
    \vspace{0pt}
\end{table}
\chen{As shown in Fig. \ref{gi_feature}, the GI exhibits more distortions but has a much larger valid area, while OptCuts on the opposite. We find the size of the valid area is more critical during the experiment, and the GI performs better with the same resolution of the feature map. Specifically, as shown in Tab. \ref{GAN_table}, our TightNet with L1 loss and geometry image for 2D mapping achieves the highest accuracy, 67.24\% for the task of tightness map prediction and 93.89\% for the task of garment segmentation. This leads to more robust multi-layer reconstruction from only a single 3D scan as following.}
% On interesting thing is that OptCut with no mask gets better IoU performance. We think that is because OptCut with no mask will predict the mask board area of the image, which contributes a lot to the IoU.

% C.1 直接无优化mesh减tightness的有毛刺的结果和我们优化的结果对比
\paragraph{Shape recovery evaluation.}
\begin{figure}[t!]
    \centering
    \includegraphics[width=0.95\linewidth]{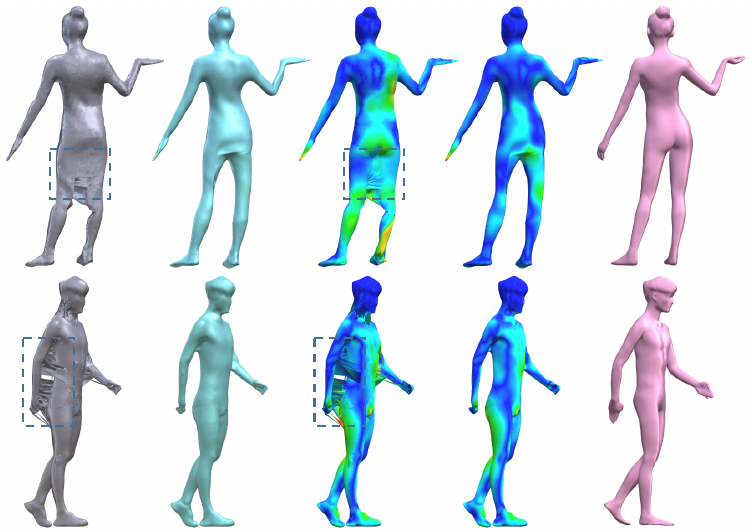}
    \caption{Evaluation of our shape recovery scheme. From left to right: the recovered body before (gray) and after (green) our shape optimization in Eq.(\ref{bodyShapeOptimization}); the corresponding per-vertex errors which are color-coded from blue to red; the ground-truth bodies (red).}
    \label{FitnessOptimization}
    \vspace{0pt}
\end{figure}
We evaluate our optimization-based shape recovery scheme by comparing it with the baseline variation using the straight-forward solution in Eq.(\ref{bodyDirect}).
As shown in Fig.\ref{FitnessOptimization}, the variation suffers from inferior reconstruction results, especially in the local regions around the oxter and crotch.
In contrast, our shape recovery scheme successfully compresses the tightness field error so as to provide accurate and visually pleasant body shape reconstruction results.

\begin{figure}[t!]
    \centering
    \includegraphics[width=0.9\linewidth]{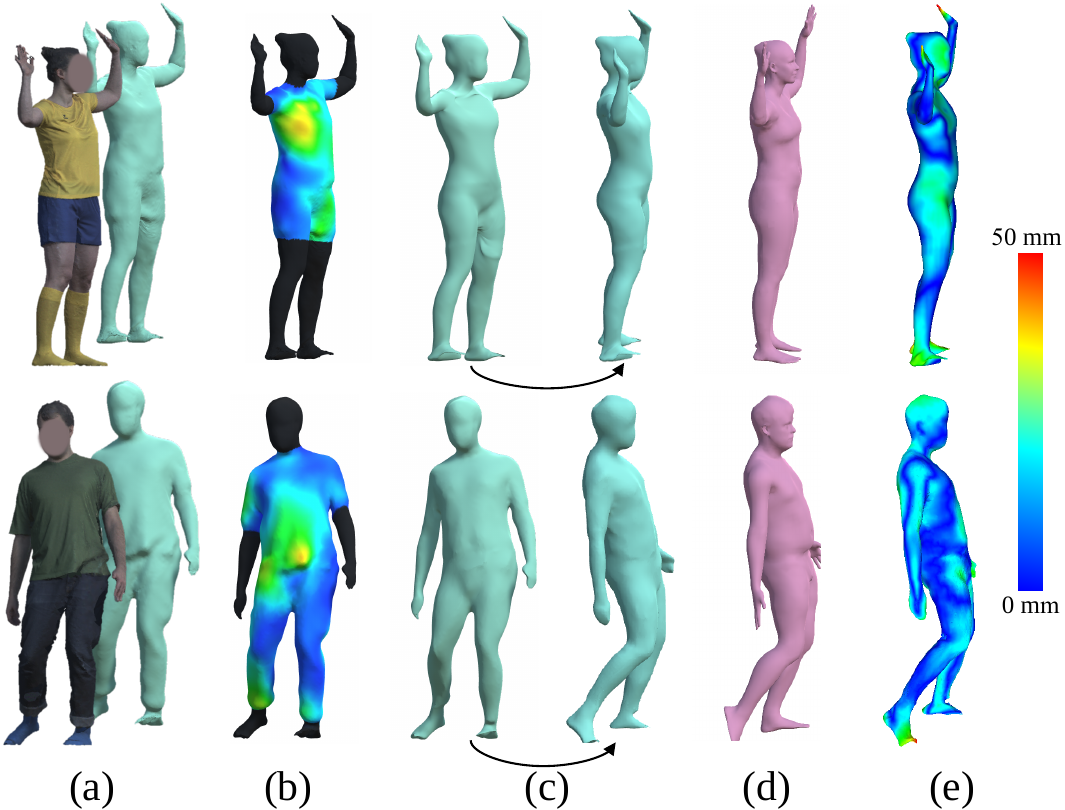}
    \caption{\revised{Comparison with our recovered body shapes and \citet{zhang2017detailed} in BUFF Dataset. (a) Input Meshes and aligned meshes. (b) Predicted tightness field on meshes. (c) The recovered body shape of our results. (d) The ground truth of \citet{zhang2017detailed}. (e) Our results with the per-vertex body shape error colored from blue (good) to red (bad).}}
    \label{shape_buff}
\end{figure}
\begin{figure}[t!]
    \centering
    \includegraphics[width=0.9\linewidth]{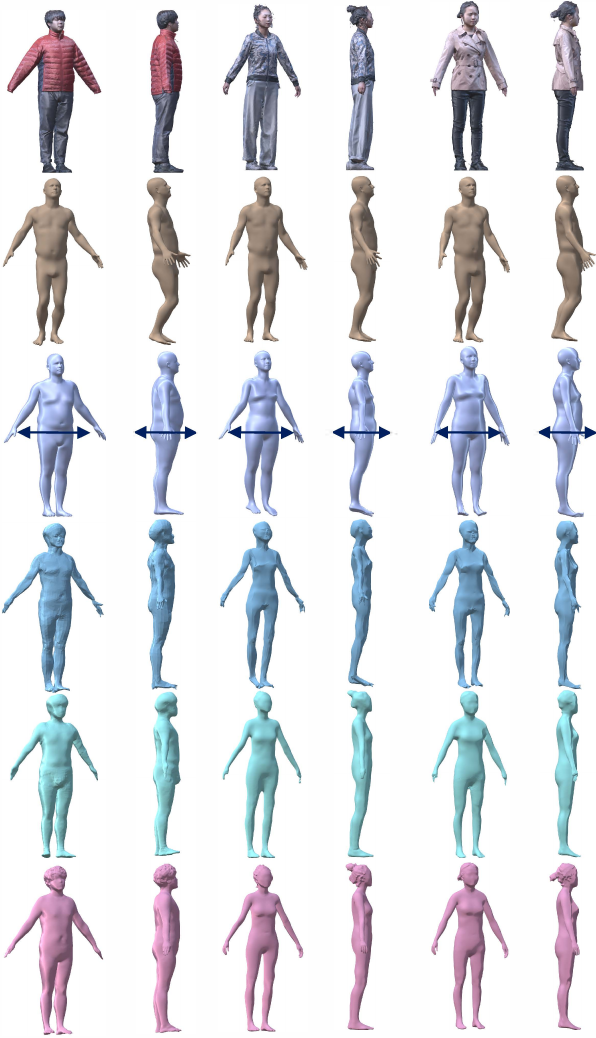}
    % \caption{\chen{The comparison of recovered body shapes in the CTD Dataset. From top to down: The input mesh; the recovered bodies using HMR~\citep{kanazawa2018end}, DoubleFusion~\citep{yu2018doublefusion}, CAPE~\citep{ma20autoenclother} and our method; the ground truth bodies. The black arrows indicate the results of DoubleFusion are always fatter with the influence of clothing.}}
    \caption{\minor{The comparison of recovered body shapes in the CTD Dataset. From top to down: The input mesh; the recovered bodies using image based method HMR~\citep{kanazawa2018end}, mesh-based method DoubleFusion~\citep{yu2018doublefusion}, GCN-based method CAPE~\citep{ma20autoenclother} and our method; the ground truth bodies. The black arrows indicate the results of DoubleFusion are always fatter with the influence of clothing.}}
    \label{Comparison_body}
\end{figure}
\revised{
\begin{table}[t!]
    \centering
    \resizebox{\linewidth}{!}{
        \begin{tabular}{l|l|c|c|c|c|c}
            \hline
            Input  & Method       & Mean$\downarrow$ & RMS$\downarrow$  & \tabincell{c}{Error                                       \\(mm)$\downarrow$} & \tabincell{c}{Front           \\IoU$\uparrow$} &\tabincell{c}{Side\\IoU$\uparrow$} \\ \hline \hline
            %\tabincell{c}{Image\\based}
            % \multirow{2}{*}{Image}
            % \multirow{Image}
            Image  & HMR          & 1.607\%          & 2.195\%          & 48.21               & 74.70\%          & 67.51\%          \\ \cline{2-7}
                   & SMPL-X       & 1.478\%          & 2.121\%          & 44.34               & 78.76\%          & 60.54\%          \\ \hline
            %\tabincell{c}{Mesh\\based}
            % \multirow{Mesh}
            A Mesh & DoubleFusion & 0.804\%          & 0.996\%          & 24.12               & 82.49\%          & 77.29\%          \\ \cline{2-7}
                   & \chen{CAPE}         & \chen{0.584\%}          & \chen{0.713\%}          & \chen{17.52}               & \chen{88.20\%}          & \chen{90.50\%}          \\ \cline{2-7}
                   & Our          & \textbf{0.191}\% & \textbf{0.451}\% & \textbf{5.73}       & \textbf{90.12}\% & \textbf{94.29}\% \\ \hline
        \end{tabular}
    }
    \caption{\chen{Comparison of recovered body shapes in CTD with HMR~\citep{kanazawa2018end}, SMPL-X~\citep{SMPLX2019}, CAPE~\citep{ma20autoenclother} and DoubleFusion~\citep{yu2018doublefusion}. \emph{Mean}, \emph{Root-Mean-Square (RMS)}, and \emph{Error (mm)} are same as Tab.~\ref{3Stage_table}. \emph{Front IoU} represents the mean IoU of each projected mask pairs (estimated body and ground-truth body) from the view of input image for HMR~\citep{kanazawa2018end} and SMPL-X~\citep{SMPLX2019}. \emph{Side IoU} uses same metric like \emph{Front IoU} but projected from the sideview.}}
    \label{shape_table}
\end{table}
}
\subsection{Comparisons of Body Recovery}
% begin
In this subsection, we demonstrate the overall performance of the proposed approach by comparing it against other state-of-the-art mesh-based and image-based body recovery methods, both qualitatively and quantitatively.

% compared methods overview
For mesh-based comparison, we compare to the state-of-the-art approaches, including the one proposed by \citet{zhang2017detailed} and the volumetric optimization stage of DoubleFusion~\citep{yu2018doublefusion}.
The former one utilizes a sequence of dressed mesh as input to recover body shape while the latter DoubleFusion~\citep{yu2018doublefusion} optimizes the body shape from a single volumetric mesh input.
For the image-based comparison, we compare to HMR~\citep{kanazawa2018end} and SMPL-X~\citep{SMPLX2019}, which regress the human model directly from only a single RGB image input.
% Comparison on BUFF, against zhang
As shown in Fig.~\ref{shape_buff}, we achieve a comparable result against the sequence-based method of \citet{zhang2017detailed} on the BUFF dataset.
Note that our method only uses a single 3D scan as input rather than a dynamic sequence of human models, which is hard to obtain for daily applications.
Besides, our network is only pre-trained using our dataset CTD without fine-tuning on BUFF, which demonstrates the generation ability for our approach to recover both human body shape and garments from only a single 3D scan.
% \paragraph{Comparison with mesh-based methods.} 
% \paragraph{Comparison with image-based methods.}

% Comparison using CTD
Then, we utilize our dataset CTD with ground-truth annotations for further qualitative and quantitative comparisons.

\definecolor{Blue}{cmyk}{1.00,1.00,0.00,0}
\revised{
\begin{table}[t!]
    \centering
    \resizebox{\linewidth}{!}{
        
        % \color{Blue}
        \begin{tabular}{l|l|c|c|c}
        
            \hline
            Input    & Method                     & \tabincell{c}{00005                                 \\T-shirt, Pants}     & \tabincell{c}{00114\\Soccer Outfit}  & Avg. \\ \hline \hline
            Image    & HMR                        & 75.08               & 44.69         & 59.89         \\ \cline{2-5}
                     & SMPL-X                     & 177.29              & 122.58        & 149.94        \\ \hline
            Mesh Seq & \citet{yang2016estimation} & 17.29               & 16.40         & 16.85         \\ \cline{2-5}
                     & detailed                   & \textbf{2.52}       & \textbf{2.23} & \textbf{2.38} \\ \hline
            A Mesh   & DoubleFusion               & 32.57               & 24.68         & 28.63         \\ \cline{2-5}
                     & Our                        & 4.73                & 3.24          & 3.98          \\ \hline
        \end{tabular}
    }
    \caption{\chen{Comparison of recovered body shapes in BUFF dataset ~\citep{zhang2017detailed} with HMR~\citep{kanazawa2018end}, SMPL-X~\citep{SMPLX2019}, DoubleFusion~\citep{yu2018doublefusion}, \citet{yang2016estimation} and detailed \citep{zhang2017detailed}. We use \emph{Root-Mean-Square (RMS)}, the same metric in the detailed \citep{zhang2017detailed}. }}
    \label{shape_table_buff}
\end{table}
}
In Fig.~\ref{Comparison_body}, we provide a qualitative comparison to DoubleFusion~\citep{yu2018doublefusion} and HMR~\citep{kanazawa2018end} on three challenging cases with various clothing tightness and similar postures to get rid of the posture ambiguity. \chen{\label{R5Q7}For a fair comparison, we also fine-tune HMR with the provided images and pose/shape parameters from optimized SMPL with ground-truth body shape.}
However, DoubleFusion~\citep{yu2018doublefusion} suffers from inferior shape recovery and turns to estimate a fatter human body without considering the clothing tightness,
while HMR~\citep{kanazawa2018end} suffers from scale ambiguity to provide only visually pleasant rather than the accurate human body shape, which is the inherent issue of such image-based methods.
In contrast, our approach accurately reconstructs the human body shape by modeling various clothing tightness field effectively.
%Although it is not fair with only an image input, this comparison can still show the precision difference in mesh/image-based methods. 

\chen{The quantitative comparison on the CTD dataset against both the mesh-based and image-based methods is provided in Tab.~\ref{shape_table}, in terms of the \emph{Mean} and \emph{Root-Mean-Square (RMS)} from Metro~\citep{cignoni1998metro} as well as the per-vertex error \emph{Error (mm)}.}
Besides, we calculate the mean IoU of each projected mask pairs (estimated body and ground-truth body) from the view of rendered image for image-based methods, denoted as \emph{Front IoU} in the CTD dataset.
We also utilize \emph{Side IoU} with the same operation as \emph{Front IoU} projected from a side-view for thorough analysis.
\chen{We also propose the comparison with CAPE~\citep{ma20autoenclother}, a representation based on graph convolutional network. We fine-tune the network of CAPE~\citep{ma20autoenclother} with aligned clothed/unclothed mesh as input/output on the CTD dataset, to utilize the GCN for body shape recovery. With the same input of a single mesh, our proposed approach utilizes higher resolution on body shape, thus produces a 1 to 2 cm decrease for the error of body shape estimation, compared with this GCN-based approach.
% With the same input of a single mesh, our proposed approach costs less computing resources and higher resolution on body shape, thus produces a 1 to 2 cm decrease for the error of body shape estimation, compared with this GCN-based approach.
}
As shown in Tab.~\ref{shape_table}, our approach achieves significantly much more accurate body recovery results in terms of all the metrics above, with the aid of modeling the influence of clothing tightness.

\chen{For the comparison on the BUFF dataset \citep{zhang2017detailed} in Tab. \ref{shape_table_buff}, our approach also achieves accurate body recovery results with only a single mesh input rather than dynamic mesh sequence. Compared to the detailed \citep{zhang2017detailed}, our approach only loses 1 to 2 millimeters but can be a more feasible approach on mobile devices.
\label{R1Q6_17}Besides, these comparisons against different inputs (images, single mesh, and mesh sequence) demonstrates that our approach not only produces the highly accurate shape but also needs a single 3D mesh, which can be more feasible for mobile devices.}

%(old version) Besides, all these comparisons against different input modularities not only highlight the robustness of our approach but also demonstrates the uniqueness of our approach as a good compromising settlement between over-demanding input data resources and highly accurate reconstruction.

% we show the recovered error between our body shapes and these ground truth body shape. For fair compassion, we compare to a 3D SMPL fitting method rather than the SMPL fitting from a single image, as our input is a 3D mesh.  This 3D SMPL fitting method use point cloud as a target to regress parameters of SMPL, which is generally used as part of human body optimization, like DoubleFusion~\citep{yu2018doublefusion}. With the help of tightness, our method can improve 0.347\% performance than these fitting methods.

\subsection{Application}\label{ClothRetargeting}
In this subsection, based on our high-quality multi-layer reconstruction, we further demonstrate various interesting applications of our approach, including immersive cloth retargeting and clothed avatar animation.

\begin{figure}[t!]
    \centering
    \includegraphics[width=0.90\linewidth]{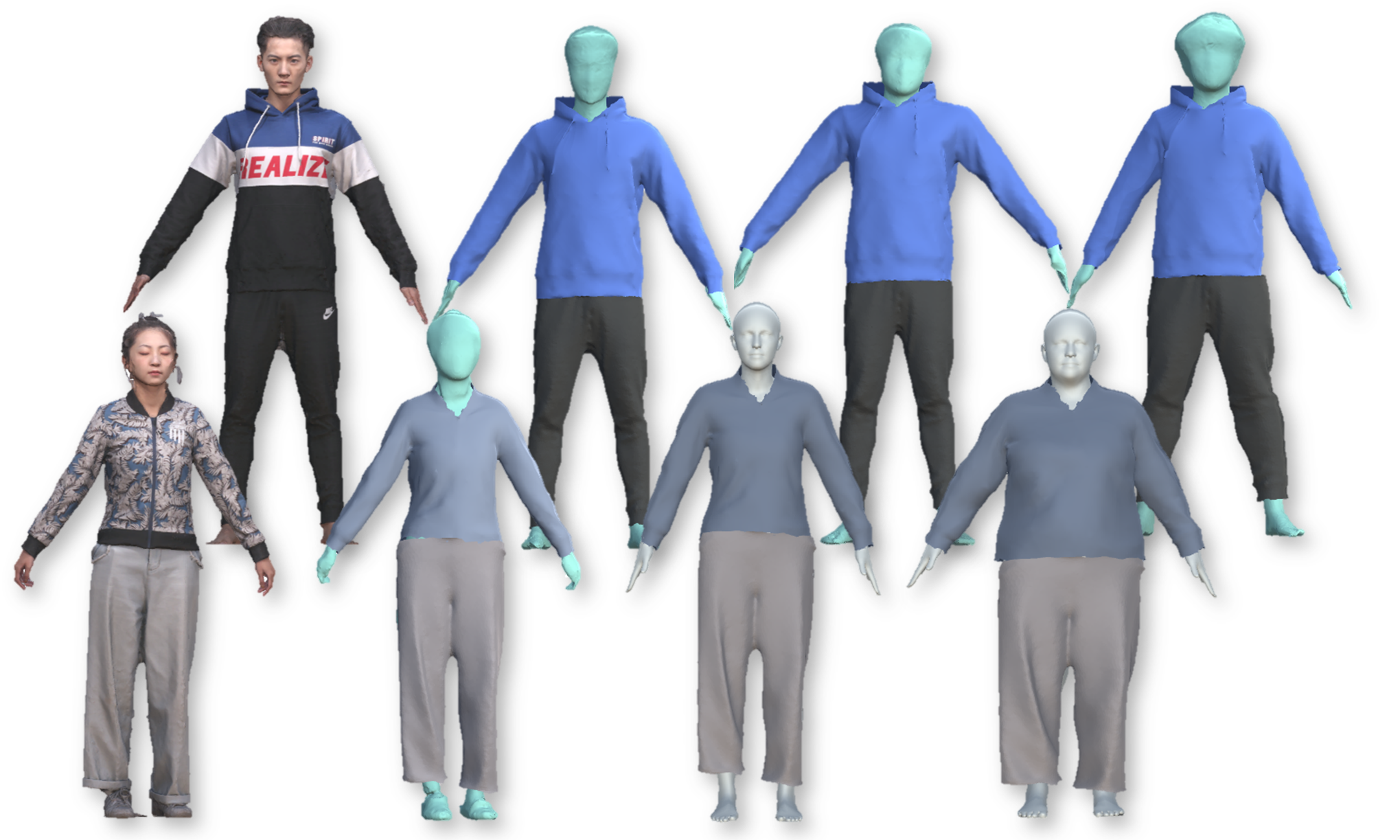}
    \caption{The application of cloth retargeting. From left to right: the input MVS meshes; the estimated body shape and segmented garments; the retargeted clothing with slim body and fat body, respectively. Note that our enhanced template models are in green while the original SMPL~\citep{loper2015smpl} models are in gray.}
    \label{fig:supp_retargeting}
\end{figure}
\begin{figure}[t!]
    \centering
    \includegraphics[width=0.85\linewidth]{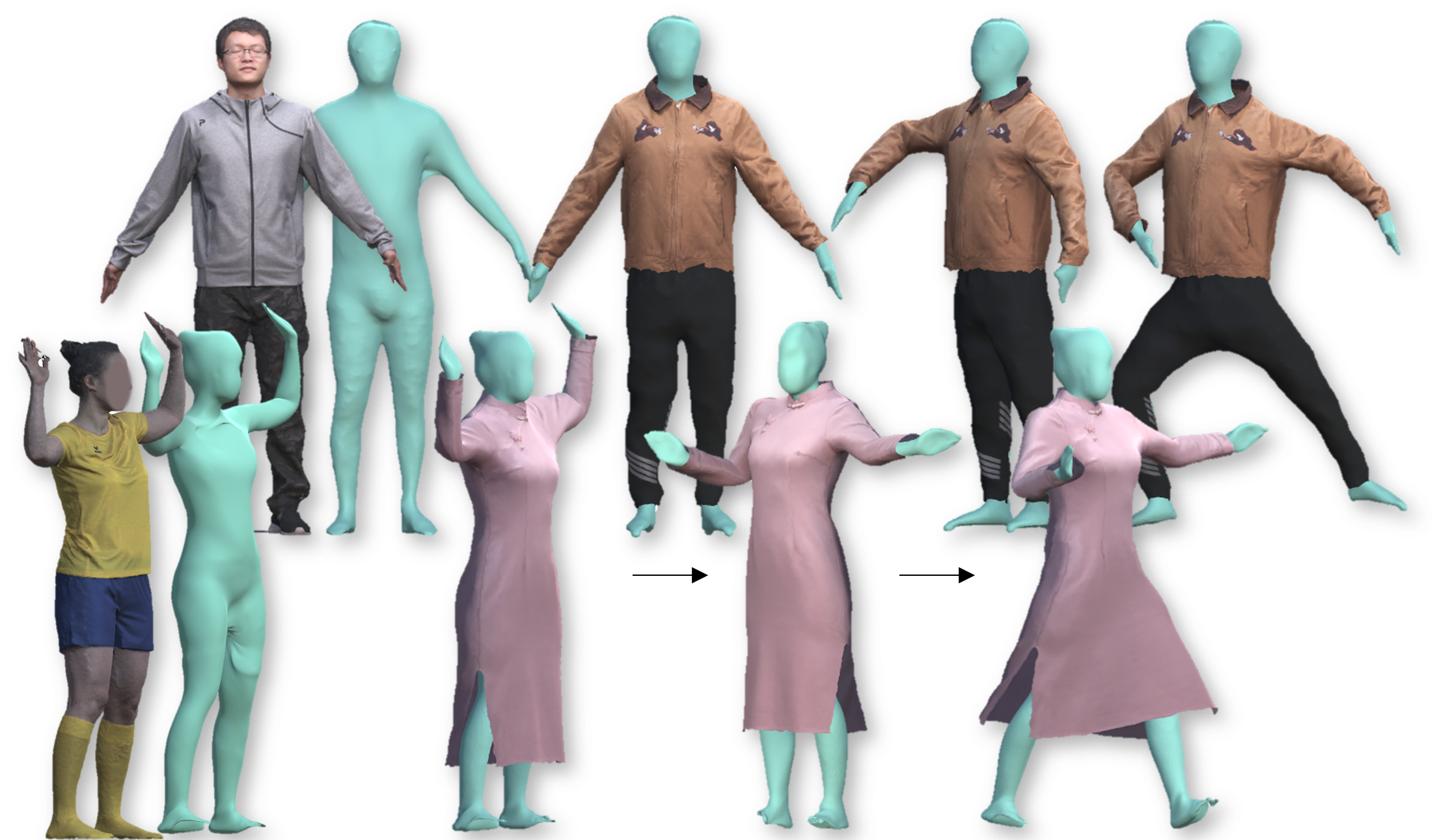}
    \caption{The application of clothed avatar animation. From left to right: The scans and our recovered inner body shape; our cloth retargeting results; our multi-layer avatar animation results into various human postures.}
    \label{fig:supp_animation}
\end{figure}

\paragraph{Cloth retargeting.}
Recall that in our approach, thanks to our novel clothing tightness fields formulation, both the aligned human template to the dressed scan in Sec.~\ref{alignment} and the recovered body shape in Sec.~\ref{ShapRecovery} share the same mesh topology and the rigged skeleton as our enhanced human template in Sec.~\ref{TemplateModel}.
Note that such displacements between the aligned template and its recovered body shape are our predicted clothing tightness.
Thus, we can directly transfer the clothing to various recovered body shapes in terms of cloth-to-body and body-to-body displacements.
As shown in Fig.\ref{fig:supp_retargeting}, we can achieve highly immersive cloth retargeting, and even fit our enhanced human template back to the original SMPL~\citep {loper2015smpl} so as to transfer the clothing to shape-variant SMPL model directly.

% We can still use this method to build the displacement between two UCBM from a different person. Hence, we can transfer clothing between different UCBM with cloth-body-body displacements. We even fit our CA-SMPL with SMPL~\citep {loper2015smpl} under different shape parameters and transfer clothing to the shape-variant SMPL model directly.

%\subsection{Clothed avatar from the static mesh}
%\label{ClothedAvatar}
\begin{figure}[t!]
    \centering
    \includegraphics[width=\linewidth]{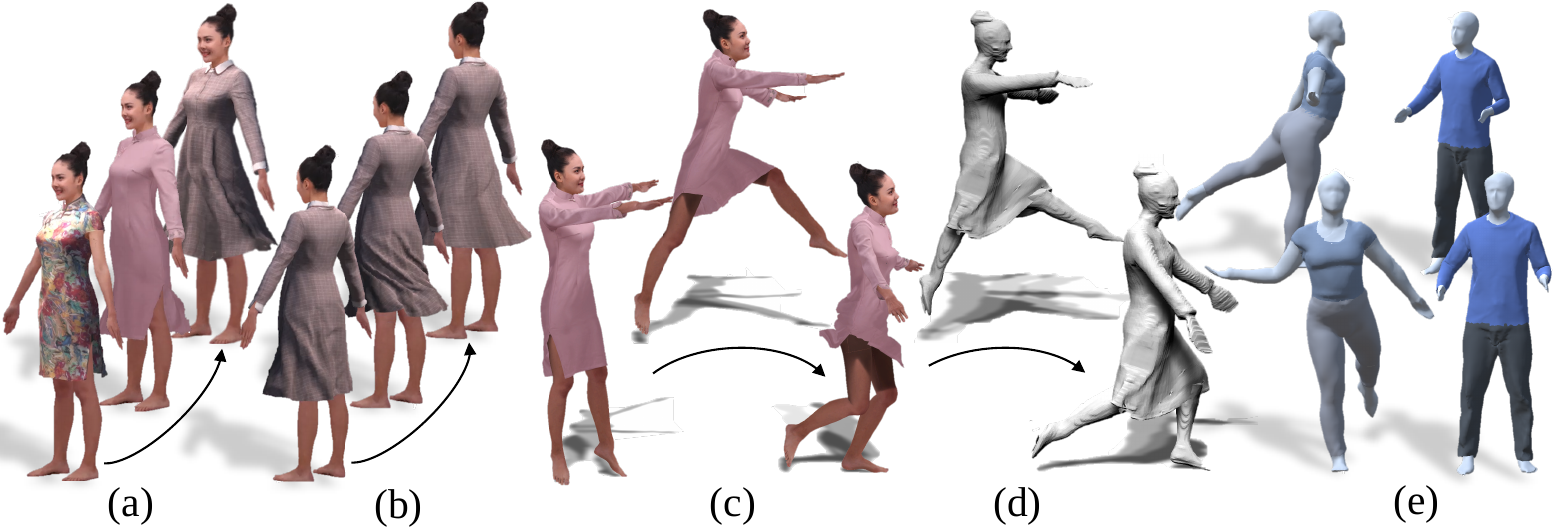}
    \caption{\chen{More results and the simulated clothes. (a) The transfered avatars with various clothes. (b) The fluttering dress with a physical simulation engine. (c) The performance from our multi-layer avatar. (d) The compared performance from the avatar of ARCH~\citet{huang2020arch}. (e) More results from the CAPE~\citep{ma20autoenclother} and MGN~\citep{bhatnagar2019mgn} dataset.} }
    \label{fig:supp_physical_animation}
\end{figure}

\paragraph{Clothed avatar animation.}
\chen{Benefiting from our enhanced human template with rigged skeleton and the novel clothing tightness field formulation, we are able to reconstruct a consistent multi-layer avatar from the input 3D scan to infer the body shape and the various segmented garments, such as the more results in Fig.~\ref{fig:supp_physical_animation} from other datasets, CAPE~\cite{ma20autoenclother} and MGN~\cite{bhatnagar2019multi}.}
Thus, we can not only change the garments of the current human target by using various clothing tightness, but also further animate the dressed avatar naturally by driving its inner body with various postures and maintaining current clothing tightness.
\chen{As shown in Fig.~\ref{fig:supp_animation} and Fig.~\ref{fig:supp_physical_animation}, we can achieve the clothed avatar animation with the rigged skeleton and support the physical simulation engine from Unity3D to generate realistic animation with our multi-layer avatars. Compared with the animated results from ARCH \citep{huang2020arch} in Fig.~\ref{fig:supp_physical_animation}, although the effect on body movement is similar, our multi-layer avatars can achieve more realistic movement like the fluttering dress.}
% xxxxx
% Note that we do not use any physical simulation for clothing, the dress floats with only skeleton-driven deformation.

\subsection{Limitations and Discussion}
\begin{figure}[t!]
    \centering
    \includegraphics[width=\linewidth]{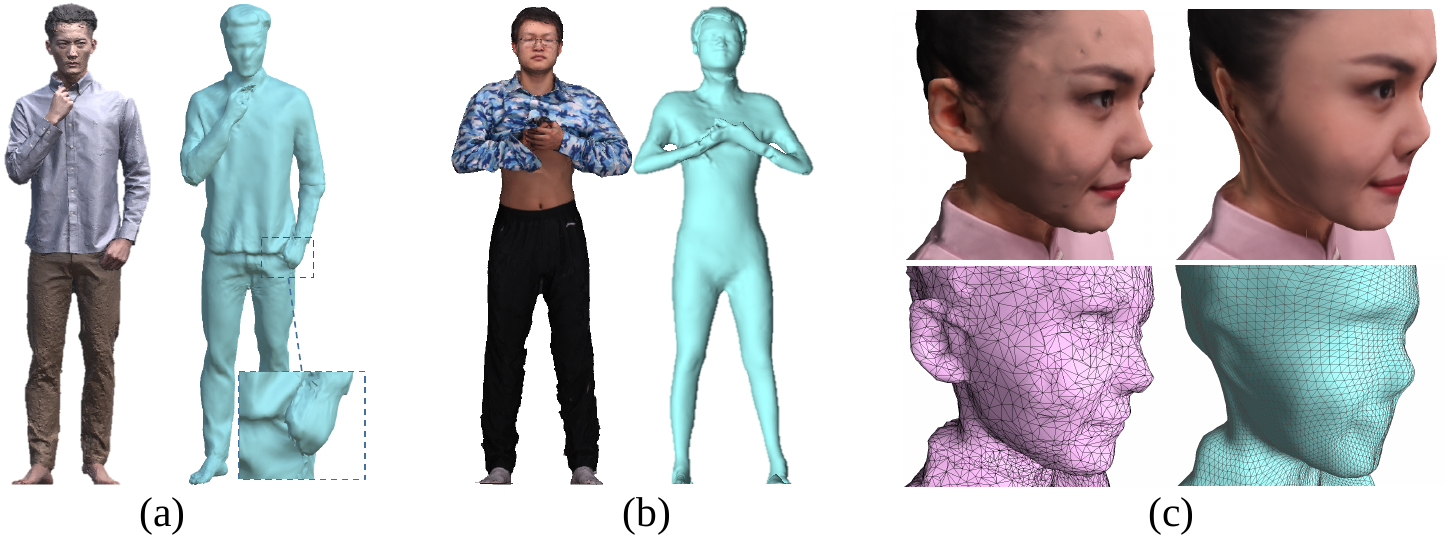}
    \caption{\chen{The failure cases of our approach. (a) The clothed human scan with the left hand in the pocket and the aligned result. (b) The clothed human scan with crossing arms, and the recovered body. 
    % \textcolor{blue}{[Ma: Could you give a description on how special the garment is. It is not clear in the figure.]}
    (c) Left column: The face of the textured scan and its mesh from the MVS approach. Right column: The face of the baked aligned mesh and its mesh from our approach.}}
    \label{fig:failurecase}
\end{figure}

Though our approach is effective for body and garment reconstruction from only a single 3D scan, it still owns limitations as follows.

\chen{
% \label{R1Q1}
First, our scheme still cannot handle extreme human poses with a merged topology such as crossing legs/arms, hidden hands or curling up, or very low-quality scans (see Fig. \ref{fig:failurecase}(b)). For those severely occluded regions such as a hand in the pocket (see Fig. \ref{fig:failurecase}(a)), our method cannot accurately deform the template as they are under-observed, which could also exist artifacts for reposing. We plan to address this by incorporating extra data-driven human hand and face key-point priors to provide a good initialization.

% \label{R5Q1}
Although our main purpose is body shape recovery and cloth modeling, the aligned region around the face and hand still needs improvement, as shown in Fig. \ref{fig:failurecase}. Without semantic processing on specific body regions, our approach still can recover fine small detail of human body parts. Thus, we plan to implement the mesh processing strategy with the semantic body part for both alignment and de-noise as one of our future works.
Besides, currently utilized geometry images in our approach can only handle genus 0 human geometry.
In reality, the human model can have a very complex topology, and a more sophisticated geometry image generation approach is required.
Alignment schemes that can handle these topologically complex human models are also our immediate future work.

% \label{R3Q4}
For specific style and of garments, like evening dress, 
% \textcolor{blue}{[Ma: Could you make a conclusion on what kinds of garments we could not handle well and the reason, like the description in the second paragraph of section 6.5? ]}, 
the proposed TightNet is still hard to generate correct results with the influence of the garment style from the dataset.
Although the TightNet is based on most effective image-to-image translation architecture, the existing seam during the mapping of geometry image still might cause the inconsistency around the boundary.
Thus, we use body shape recovery function to smooth between these regions.

% \label{R1Q4}
For our applications, although we can achieve both skeleton-driven and physical simulated clothed avatar animation in Fig.~\ref{fig:supp_animation} and Fig.~\ref{fig:supp_physical_animation}, our current clothing tightness field formulation still cannot simulate the dynamic movement of clothing in a physically plausible manner for our reconstructed garments. 
It's a promising direction to further model the clothing tightness field for 4D dynamic sequences with the aid of extra-human motion priors like CAPE \citep{ma20autoenclother} and \citet{ZhouWLCYSLS20}.
Moreover, our approach relies on raw 3D human scans, which are usually difficult to obtain, and the quality can not be guaranteed.
Hence we plan to explore the possibility of directly taking a single or sparse set of 2D images \citep{chen2018autosweep} as the input of the MVS setting for recovering the 3D clothing and human shape.
Also, through augmented training under various lighting conditions using the light stage, it is possible to capture the reflection property of the clothing and for a better AR/VR or try-on experience.
}

%%%%%%%%%%%%%%%%%%%%%%%%%%%%%%%%%%%%%%%%%%%%%%%%%%%%%%%%%%%%%%%%%%%%%%%%%%%%%%%%%%%%%%%%%%%%%%%%%%%%%%%%%%%%%%%%%%%%%%%%
\section{Conclusion}
We present TightCap, a learning-based scheme for robustly and accurately capturing the clothing tightness field as well as human geometry with a single clothed 3D human raw mesh.
The key contribution of our approach is the usage of geometry image for tightness prediction, and the alignment of human geometry enables the geometry image correspondence from various types of clothing.
Moreover, we collect a large 3D Clothing Tightness Dataset (CTD) for the clothed human reconstruction tasks.
We propose and train a modified conditional GAN network to automatically predict the clothing tightness map and, subsequently, the underlying human shape.
Experiments demonstrate the reliability and accuracy of our method.
We also exhibit two interesting virtual try-on applications, i.e., cloth retargeting and clothed avatar animation.
We believe our scheme will benefit various AR/VR research and applications, such as virtual try-on and avatar animation.

% 不make sense，加lighting为什么会影响tightness？
% By augmenting the training with lighting estimation, it may be possible to directly clothing tightness from a single 2D image.
%一直在重复
%In summary, our method demonstrates that tightness prediction is a feasible method for recovering human body shape and segmenting cloth with only one static input mesh. We also collect a large dataset (CTD) with variant body shapes and clothes to help the researches of body estimation or clothing segmentation. We believe this learning-based scheme for body shape will enable body estimation and virtual-try with only a static mesh captured from a depth sensor of mobile phone and will help more applications in VR/AR.
% We have presented the first data-driven method that estimates clothing tightness considering clothing type and geometry as well as human pose and is capable of recovering human body shape and segmenting cloth. We rely on a clothing-aware template model modified from SMPL. Given a single clothed human mesh, we align it with our template model, map its key geometry into clothed-GI, estimate its tightness map, recover body shape, and segment clothing. 
%%%%%%%%%%%%%%%%%%%%%%%%%%%%%%%%%%%%%%%%%%%%%%%%%%%%%%%%%%%%%%%%%%%%%%%%%%%%%%%%%%%%%%%%%%%%%%%%%%%%%%%%%%%%%%%%%%%%%%%%
%\end{CJK}

\begin{acks}
	The authors would like to thank WenGuang Ma, YeCheng Qiu, MingGuang Chen for help with data acquisition; Hongbo Wang, Gao Ya, Shenze Ye, Teng Su for help with data annotation. This work is supported by the National Key Research and Development Program (2018YFB2100500), the programs of NSFC (61976138 and 61977047), STCSM (2015F0203-000-06), and SHMEC (2019-01-07-00-01-E00003).
\end{acks}

\bibliographystyle{ACM-Reference-Format}
\bibliography{sample-base}

%\appendix









\end{document}